\documentclass[journal,transmag,onecolumn]{IEEEtran}
\IEEEoverridecommandlockouts

\usepackage{bm,mathtools,amsfonts,amssymb,bbm,float,hyperref,euscript,setspace,tikz,steinmetz,cite,wrapfig,subcaption}

\usepackage{lastpage}
\usepackage{amsmath}

\usepackage{enumitem}

\usepackage[normalem]{ulem}
\usepackage[mode=buildnew]{standalone}
\usepackage{mathabx} 
\newtheorem{theorem}{Theorem}
\newtheorem{lemma}{Lemma}

\newtheorem{definition}{Definition}

\newtheorem{corollary}{Corollary}

\newtheorem{assumption}[theorem]{Assumption}
\newtheorem{proposition}{Proposition}

\usepackage{mleftright}\mleftright

\definecolor{darkgreen}{rgb}{0, 0.5, 0}
\definecolor{darkred}{RGB}{128, 0, 0}

\newcommand{\stkout}[1]{\ifmmode\text{\sout{\ensuremath{#1}}}\else\sout{#1}\fi}
\newcommand{\markov}{\mathrel{\multimap}\joinrel\mathrel{-}%
	\joinrel\mathrel{\mkern-6mu}\joinrel\mathrel{-}}

\newcommand{\vc}[1]{\mathbf{#1}} 
\newcommand{\der}{\mathrm{d}} %

\newcommand{\ra}{\rightarrow}
\newcommand{\eg}{\emph{e.g., }}
\newcommand{\ie}{\emph{i.e., }}

\DeclareMathOperator{\gen}{gen}

\DeclareMathOperator{\supp}{supp}

\DeclareMathOperator*{\argmin}{arg\,min}

\newcommand{\T}{\mathsf{T}}
\newcommand{\dT}{\Delta t}

\newcommand{\iid}{i.i.d. }

\makeatletter
\DeclareFontFamily{OMX}{MnSymbolE}{}
\DeclareSymbolFont{MnLargeSymbols}{OMX}{MnSymbolE}{m}{n}
\SetSymbolFont{MnLargeSymbols}{bold}{OMX}{MnSymbolE}{b}{n}
\DeclareFontShape{OMX}{MnSymbolE}{m}{n}{
	<-6>  MnSymbolE5
	<6-7>  MnSymbolE6
	<7-8>  MnSymbolE7
	<8-9>  MnSymbolE8
	<9-10> MnSymbolE9
	<10-12> MnSymbolE10
	<12->   MnSymbolE12
}{}
\DeclareFontShape{OMX}{MnSymbolE}{b}{n}{
	<-6>  MnSymbolE-Bold5
	<6-7>  MnSymbolE-Bold6
	<7-8>  MnSymbolE-Bold7
	<8-9>  MnSymbolE-Bold8
	<9-10> MnSymbolE-Bold9
	<10-12> MnSymbolE-Bold10
	<12->   MnSymbolE-Bold12
}{}

\let\llangle\@undefined
\let\rrangle\@undefined
\DeclareMathDelimiter{\llangle}{\mathopen}%
{MnLargeSymbols}{'164}{MnLargeSymbols}{'164}
\DeclareMathDelimiter{\rrangle}{\mathclose}%
{MnLargeSymbols}{'171}{MnLargeSymbols}{'171}
\makeatother

\title{Data-dependent Generalization Bounds \\ via Variable-Size Compressibility}

\author{Milad Sefidgaran and Abdellatif Zaidi%
\thanks{The material of this paper has been published partially at the 2024 International Zurich Seminar on Information and Communication (IZS) and partially at the 2024 IEEE International Symposium of Information Theory (ISIT).}
\thanks{M. Sefidgaran is with the Paris Research Center of Huawei Technologies France,  e-mail: milad.sefidgaran2@huawei.com.}%
\thanks{A. Zaidi is with the Universit\'e Gustave Eiffel and the Paris Research Center of Huawei Technologies France, e-mail: abdellatif.zaidi@univ-eiffel.fr.}
}

\begin{document}
\maketitle


\begin{abstract}%
	In this paper, we establish novel \textit{data-dependent} upper bounds on the generalization error through the lens of a ``variable-size compressibility'' framework that we introduce newly here. In this framework, the generalization error of an algorithm is linked to a variable-size `compression rate' of its input data. This is shown to yield bounds that depend on the empirical measure of the given input data at hand, rather than its unknown distribution. Our new generalization bounds that we establish are tail bounds, tail bounds on the expectation, and in-expectations bounds. Moreover, it is shown that our framework also allows to derive general bounds on \textit{any} function of the input data and output hypothesis random variables. In particular, these general bounds are shown to subsume and possibly improve over several existing PAC-Bayes and data-dependent intrinsic dimension-based bounds that are recovered as special cases, thus unveiling a unifying character of our approach. For instance, a new data-dependent intrinsic dimension-based bound is established, which connects the generalization error to the optimization trajectories and reveals various interesting connections with the rate-distortion dimension of a process, the Rényi information dimension of a process, and the metric mean dimension.
\end{abstract}




\section{Introduction and problem setup}
Let $Z\in \mathcal{Z}$ be some \emph{input data} distributed according to an unknown distribution $\mu$, where $\mathcal{Z}$ is the \emph{data space}. A major problem in statistical learning is to find a \emph{hypothesis} (model) $w$ in the \emph{hypothesis space} $\mathcal{W}$ that minimizes the \emph{population risk} defined as \cite{shalev2014understanding}
\begin{align}
	\mathcal{L}(w)\coloneqq \mathbb{E}_{Z \sim \mu}\left[\ell(Z,w)\right],\quad w \in \mathcal{W},
	\label{def:popRisk}
\end{align}
where $\ell:\mathcal{Z} \times \mathcal{W}\to \mathbb{R}^+$ is a loss function that measures the quality of the prediction of the hypothesis $w \in \mathcal{W}$. The distribution $\mu$ is assumed to be unknown, however; and one has only access to $n$ (training) samples $S=\{Z_1,\ldots,Z_n\}\sim P_{S}=\mu^{\otimes n}$ of the input data. Let $\mathcal{A} \colon \mathcal{S} \ra \mathcal{W}$, $\mathcal{S}=\mathcal{Z}^n$, be a possibly stochastic algorithm which, for a given input data $s=\{z_1,\hdots,z_n\} \in \mathcal{Z}^n$, picks the hypothesis $\mathcal{A}(s) = W \in \mathcal{W}$. This induces a conditional distribution $P_{W|S}$ on the hypothesis space $\mathcal{W}$. Instead of the population risk minimization problem~\eqref{def:popRisk} one can consider minimizing the \emph{empirical risk}, given by
\begin{align}
	\mathcal{\hat{L}}(s,w)\coloneqq \frac{1}{n}\sum\nolimits _{i=1}^m \ell(z_i,w). 
	\label{def:empRisk}
\end{align}
Nonetheless, the minimization of the empirical risk (or a regularized version of it) is meaningful only if the difference between the population and empirical risks is small enough. This difference is known as the \emph{generalization error} of the learning algorithm and is given by
\begin{align}
	\gen(s,\mathcal{A}(s)) \coloneqq \mathcal{L}(\mathcal{A}(s)) -	\mathcal{\hat{L}}(s,\mathcal{A}(s)). 
	\label{def:gen}
\end{align}
An exact analysis of the statistical properties of the generalization error~\eqref{def:gen} is out-of-reach, however, except in very few special cases; and, often, one resorts to bounding the generalization error from the above, instead. The last two decades have witnessed the development of various such upper bounds, from different perspectives and by undertaking approaches that often appear unrelated. Common approaches include information-theoretic, compression-based, fractal-based, or intrinsic-dimension-based, and PAC-Bayes ones. Initiated by Russo
and Zou~\cite{russozou16}and Xu and Raginsky~\cite{xu2017information}, the information-theoretic approach measures the complexity of the hypothesis space by the Shannon mutual information between the input data and the algorithm output. See also the follow-up works~\cite{steinke2020reasoning,esposito2020, Bu2020,haghifam2021towards,neu2021informationtheoretic,aminian2021exact,Zhou2022,lugosi2022generalization,masiha2023f}. The roots of compression-based approaches perhaps date back to Littlestone and Warmuth~\cite{littlestone1986relating} who studied the predictability of the training data labels using only part of the dataset. This compressibility approach has been extended in various ways in several subsequent works, including to elaborate \emph{data-dependent} bounds such as in~\cite{hanneke2019sharp,hanneke2019sample,bousquet2020proper,hanneke2021stable,hanneke2020universal} and~\cite{cohen2022learning}. Closer to our work is another popular compressibility approach that studies the compressibility of the \emph{hypothesis space}, see, e.g.,~\cite{arora2018stronger, suzuki2018spectral,hsu2021generalization,barsbey2021heavy} and the recent \cite{Sefidgaran2022}. The fractal-based approach is a recently initiated line of work that hinges on that when the algorithm has a recursive nature, e.g., it involves an iterative optimization procedure, it might generate a fractal structure either in the model trajectories~\cite{simsekli2020hausdorff,birdal2021intrinsic,hodgkinson2022generalization,lim2022chaotic} or in its distribution~\cite{camuto2021fractal}. These works show that, in that case, the generalization error is controlled by the intrinsic dimension of the generated fractal structure. The original PAC-Bayes bounds were stated for classification~\cite{mcallester1998some,mcallester1999pac}; and, it has then become clear that the results could be extended to any bounded loss, resulting in many variants and extensions of them~\cite{seeger2002pac,langford2001not,catoni2003pac,maurer2004note,germain2009pac,tolstikhin2013pac,begin2016pac,thiemann2017strongly,dziugaite2017computing,neyshabur2018pacbayesian,rivasplata2020pac,negrea2020defense,negrea2020it,viallard2021general}. For more details on PAC-Bayes bounds, we refer the reader to Cantoni's book~\cite{catoni2007pac} or the tutorial paper of~\cite{alquier2021}.  

The aforementioned approaches have evolved independently of each other; and the bounds obtained with them differ in many ways that it is generally difficult to compare them.\footnote{Two months after we submitted the initial version of this work on arXiv, Chu \& Raginsky \cite{chu2023unified} proposed an alternative way to unifying information-theoretic and PAC-Bayes bounds. For these two types of bounds, their approach has the advantage of being technically simpler comparatively but it leads to bounds that only hold true up to some constants, in contrast with ours which recovers them optimally. Chu-Raginsky approach however does not recover dimension-based and rate-distortion-theoretic based bounds.} Arguably, however, the most useful bounds must be \textit{computable}. This means that the bound should depend on the particular sample of the input data at hand, rather than just on the distribution of the data which is unknown. Such bounds are called~\textit{data-dependent}; they are preferred and are generally of bigger utility in practice. In this sense, most existing information-theoretic and rate-distortion theoretic-based bounds on the generalization error are data-independent. This includes the mutual information bounds of Russo and Zou~\cite{russozou16} and Xu and Raginsky~\cite{xu2017information} whose computation requires knowledge of the joint distribution of the input data and output hypothesis; and, as such, they are not computable with just one sample training dataset at hand. In fact, the most prominent \emph{data-dependent} bounds are those obtained with the PAC-Bayes approach. Note that although their numerical evaluation may not be straightforward in general, especially for large neural networks, a few works have leveraged the compressibility/pruning principle to compute the PAC-Bayes bounds~\cite{zhou2018non,lotfi2022pac}.

\textbf{Contributions.} In this paper, we establish novel \textit{data-dependent} generalization bounds through the lens of a ``variable-size compressibility'' framework that we introduce here. In this framework, the generalization error of an algorithm is linked to a variable-size `compression rate' of the input data. This allows us to derive bounds that depend on the particular empirical measure of the input data, rather than its unknown distribution. The novel generalization bounds that we establish are tail bounds, tail bounds on the expectation, and in-expectations bounds. Moreover, we show that our variable-size compressibility approach is somewhat generic and it can be used to derive general bounds on \textit{any} function of the input data and output hypothesis random variables -- In particular, see our general tail bound of Theorem~\ref{th:generalTail} in Section~\ref{sec:generalTail} as well as other bounds in that section. In fact, as we show, the framework can be accommodated easily to encompass various forms of tail bounds, tail bounds on the expectation, and in-expectation bounds through judicious choices of the distortion measure. In particular, in Section~\ref{sec:applications}, by specializing them we show that our general variable-size compressibility bounds subsume various existing data-dependent PAC-Bayes and intrinsic-dimension-based bounds and recover them as special cases. Hence, another advantage of our approach is that it builds a unifying framework that allows formal connections with the aforementioned, seemingly unrelated, Rate-distortion theoretic, PAC-Bayes, and dimension-based approaches.

In some cases, our bounds are shown to be novel and possibly tighter than the existing ones. For example, see Proposition~\ref{prop:PAClossy} for an example new data-dependent PAC-Bayes type bound that we obtain readily by application of our general bounds and some judicious choices of the involved variables. Also, our Theorem~\ref{th:RDProcess} provides a new dimension-based type bound, which states that the average generalization error over the trajectories of a given algorithm can be upper bounded in terms of the amount of compressibility of such optimization trajectories, measured in terms of a suitable rate-distortion function. This unveils novel connections between the generalization error and newly considered data-dependent intrinsic dimensions, including the rate-distortion dimension of a process, the metric mean dimension, and the Rényi information dimension of a process. The latter is sometimes used in the related literature to characterize the fundamental limits of compressed sensing algorithms. We emphasize that the key ingredient to our approach is the new ``variable-size compressibility'' framework that we introduce here. For instance, the framework of \cite{Sefidgaran2022}, which comparatively can be thought of as being of ``fixed-size'' compressibility type, only allows to derive data-independent bounds; and, so, it falls short of establishing any meaningful connections with PAC-Bayes and data-dependent intrinsic dimension-based bounds. This is also reflected in the proof techniques in our case which are different from those for the fixed-size compressibility framework. 

The rest of this paper is organized as follows. In Section~\ref{sec:compress}, we introduce our variable-size compressibility framework and provide a data-dependent tail bound on the generalization error. Section~\ref{sec:generalBounds} contains our general bounds. Section~\ref{sec:applications} provides various applications of our main results, in particular, to establish PAC-Bayes and dimension-based bounds which are also verified by experiments in Section~\ref{sec:experiments}. The proofs, as well as other related results, are deferred to the appendices.

\paragraph{Notations.} We denote random variables, their realizations, and their alphabets by upper-case letters, lower-case letters, and calligraphy fonts; \eg $X$, $x$, and $\mathcal{X}$. The distribution, the expected value, and the support set of a random variable $X$ are denoted as $P_X$, $\mathbb{E}[X]$, and $\supp(P_X)$. A random variable $X$ is called $\sigma$-subgaussian, if $\log \mathbb{E}[\exp(\lambda (X-\mathbb{E}[X]))]\leq \lambda^2 \sigma^2/2$, $\forall \lambda \in \mathbb{R}$.\footnote{Throughout, $\log(\cdot)$ designates the natural logarithms.} A collection of $m\in \mathbb{N}$ random variables $(X_1,\ldots,X_m)$ is denoted as $X^m$ or $\vc{X}$, when $m$ is known by the context. The notation $\{x_i\}_{i=1}^m$ is used to represent $m$ real numbers; used also similarly for sets or functions. We use the shorthand notation $[m]$ to denote integer ranges from $1$ to $m \in \mathbb{N}$. Finally, the non-negative real numbers are denoted by $\mathbb{R}^+$.

Most of our results are expressed in terms of information-theoretic functions.\footnote{The reader is referred to~\cite{CoverTho06,elgamal_kim_2011,CsisKor82,polyanskiy2014lecture} for further details on this.} For two distributions $P$ and $Q$ defined on the same measurable space, define the \emph{Rényi divergence} of order $\alpha\in \mathbb{R}$, when $\alpha\neq 1$, as $D_{\alpha}\left(Q\|P\right) \coloneqq \frac{1}{\alpha-1}\log \mathbb{E}_Q\left[\left(\frac{\der Q}{\der P}\right)^{\alpha-1}\right]$, if $Q \ll P$, and $\infty$ otherwise. Here, $\frac{\der Q}{\der P}$ is the Radon-Nikodym derivative of $Q$ with respect to $P$. The \emph{Kullback–Leibler} (KL) divergence is defined as $D_{KL}\left(Q\|P\right) \coloneqq \mathbb{E}_Q\left[\log\frac{\der Q}{\der P}\right]$, if $Q \ll P$, and $\infty$ otherwise. Note that $\lim_{\alpha \to 1} D_{\alpha}(Q\|P)=D_{KL}(Q\|P)$. The \emph{mutual information} between two random variables $X$ and $Y$, distributed jointly according to $P_{X,Y}$ with marginal $P_X$ and $P_Y$, is defined as $I(X;Y)\coloneqq D_{KL}\left(P_{X,Y}\|P_X P_Y\right)$. This quantity somehow measures the \emph{amount} of information $X$ has about $Y$, and vice versa. Finally, throughout we will make extensive usage of the following sets, defined for a random variable $X \in \mathcal{X}$ with distribution $P_X$ and a real-valued function $g(X)\colon \mathcal{X} \to \mathbb{R}$ as
\begin{align}
	\mathcal{G}_{X}^{\delta} \coloneqq& \left\{\nu_{X} \in \mathcal{P}_{\mathcal{X}} \colon D_{KL}\left(\nu_{X} \| P_{X}\right)\leq \log(1/\delta)\right\}, \label{def:Gdelta}\\
	\mathcal{S}_{X}\left(g(X)\right) \coloneqq& \left\{\nu_{X} \in \mathcal{P}_{\mathcal{X}} \,\big| \, \forall x \in \supp(\nu_X) \colon g(x)> 0\right\}, \label{def:fSupport}
\end{align}
where $\mathcal{P}_{\mathcal{X}}$ is the set of all distributions defined over $\mathcal{X}$. 


\section{Variable-size compressibility} \label{sec:compress}

As we already mentioned, the approach of Sefidgaran et al.~\cite{Sefidgaran2022} is based on a fixed-size compressibility framework; and, for this reason, it only accommodates bounds on the generalization error that are independent of the data. In this work, we develop a ``variable-size'' compressibility framework, which is more general and allows us to establish new data-dependent bounds on the generalization error. As it will become clearer throughout, in particular, this allows us to build formal connections with seemingly-unrelated approaches such as PAC-Bayes and data-dependent intrinsic dimension bounds. 

We start by recalling the aforementioned fixed-size compressibility framework, which itself can be seen as an extension of the classic compressibility framework found in source coding literature. We refer the reader to~\cite{CoverTho06} for an easy introduction to classic rate-distortion theory. It should be noted that the rate-distortion theory was also used in \cite{Bu2021ModelCompression} to study how the generalization error of the ``compressed model'' deviates from that of the original (uncompressed) model. Here, similar to \cite{Sefidgaran2022}, we use the rate-distortion theory to bound the generalization error of the original model.

Consider a learning algorithm $\mathcal{A}(S)\colon \mathcal{S}\to \mathcal{W}$. The goal of the compression for the generalization error problem is to find a suitable \emph{compressed} learning algorithm $\mathcal{\hat{A}}(S,W)=\hat{W} \in \mathcal{\hat{W}}\subseteq \mathcal{W}$ which has a smaller \textit{complexity} than that of the original algorithm $\mathcal{A}(S)$ and whose generalization error is close enough to that of $\mathcal{A}(S)$. Define the distortion function $\tilde{d} \coloneqq \mathcal{S} \times \mathcal{W} \times \mathcal{\hat{W}} \to \mathbb{R}$ as $\tilde{d}(w,\hat{w};s) \coloneqq \gen(s,w)-\gen(s,\hat{w})$. In order to guarantee that  $\tilde{d}(\mathcal{A}(S),\mathcal{\hat{A}}(S,\mathcal{A}(S)))$ does not exceed some desired threshold one needs to consider the \emph{worst-case} scenario; and, in general, this results in looser bounds. Instead, they considered an adaptation of the \emph{block-coding} technique, previously introduced in the source coding literature, for the learning algorithms. Consider a block of $m\in \mathbb{N}$ datasets $s^m=(s_1,\ldots,s_m)$ and one realization of the associated hypotheses $w^m=(w_1,\ldots,w_m)$, with $w_i=\mathcal{A}(s_i)$ for $i\in [m]$, which we denote in the rest of this paper with a slight abuse of notation as $\mathcal{A}(s^m)=w^m$. In this technique, the compressed learning algorithm $\mathcal{A}(s^m,w^m)\colon \mathcal{S}^m \times \mathcal{W}^m \to \mathcal{\hat{W}}^m$ is allowed to \emph{jointly} compress these $m$ instances, to produce $\mathcal{\hat{W}}^m$. Let, for given $s^m$ the distortion between the output hypothesis of algorithm $\mathcal{A}(\cdot)$ applied on the vector $s^m$, i.e., $w^m=\mathcal{A}(s^m)$, and its compressed version $\hat{A}(\cdot,\cdot)$ applied on the vector $(s^m,w^m)$, i.e., $\hat{w}^m = \hat{\mathcal{A}}(s^m,w^m)$, be the average of the element-wise distortions $\tilde{d}(\cdot,\cdot,\cdot)$ between their components, $\tilde{d}_m(w^m,\hat{w}^m;s^m) \coloneqq \frac{1}{m}\sum\nolimits_{i\in [m]} \tilde{d}(w_i,\hat{w}_i;s_i)$. As is easily seen, this block-coding approach with average distortion enables possibly smaller distortion levels, in comparison with those allowed by worst-case distortion over the components.

Sefidgaran et al.~introduced the following definition of (exponential) compressibility, which they then used to establish \textit{data-independent} tail and in-expectation bounds on the generalization error. Denote by $\mathbb{P}_{(S,W)^{\otimes m}}$ the probability with respect to the $m$-times product measure of the joint distribution of $S$ and $W$.
\begin{definition}[{\cite[Definition~8]{Sefidgaran2022}}] \label{def:compressOld} The learning algorithm \,$\mathcal{A}$ is called $(R,\epsilon,\delta;\tilde{d}_m)$-compressible\footnote{For convenience, we drop out the dependence on $\mu$, $n$, and $P_{W|S}$ in the definition.} for some $R,\delta \in \mathbb{R}^+$ and $\epsilon \in \mathbb{R}$, if there exists a sequence of hypothesis books $\{\mathcal{H}_m\}_{m \in \mathbb{N}}$, $\mathcal{H}_m=\{\hat{\vc{w}}[j],j\in [l_m]\} \subseteq \mathcal{\hat{W}}^m$ such that $l_m \leq e^{mR}$ and
	\begin{align}
		\lim_{m\to \infty} \bigg[-\frac{1}{m}\log\mathbb{P}_{(S,W)^{\otimes m}}\Big(\min_{j \in [l_m]} \tilde{d}_m(W^m,\hat{\vc{w}}[j];S^m) >\epsilon \Big)\bigg]\geq \log(1/\delta). \label{eq:oldComp}
	\end{align}	
\end{definition}
\noindent The inequality~\eqref{eq:oldComp} expresses the condition that, for large $m$, the probability (over $(S^m, W^m)$) of finding \textit{no} $\hat{\mathbf{w}}[j]$ that is within a distance less than $\epsilon$ from $W^m$ vanishes faster than $\delta^m$. Equivalently, the probability that the distance from $W^m$ of any element $\hat{\mathbf{w}}[j]$ of the book exceeds $\epsilon$ (sometimes called probability of ``excess distortion'' or ``covering failure'') is smaller than $\delta^m$ for large $m$.

A result of~\cite[Theorem~9]{Sefidgaran2022} states that if $\mathcal{A}$ is $(R,\epsilon,\delta;\tilde{d}_m)$-compressible in the sense of Definition~\ref{def:compressOld} and the loss $\ell(Z,w)$ is $\sigma$-subgaussian for every $w \in \mathcal{W}$ then with probability $(1-\delta)$ it holds that 
\begin{equation}
	\gen(S,W)\leq \sqrt{2\sigma^2(R+\log(1/\delta))/n}+\epsilon.
	\label{tail-bound-Sefidgaran-et-al-theorem9}
\end{equation}
Also, let $R(\delta,\epsilon)\coloneqq \sup_{Q\in \mathcal{G}_{S,W}^{\delta}} \mathfrak{RD}(\epsilon;Q)$ where
\begin{align}
	\mathfrak{RD}(\epsilon;Q)\coloneqq \inf\nolimits _{P_{\hat{W}|S}} I(S;\hat{W}),\quad\text{s.t.}\quad \mathbb{E}[\gen(S,W)-\gen(S,\hat{W})]\leq \epsilon,
	\label{rate-distortion-function-generic-distribution-Q}
\end{align}
the supremum is over all distributions $Q$ over $\mathcal{S} \times \mathcal{W}$ that are in the $\delta$-vicinity of the joint $P_{S,W}$ in the sense of~\eqref{def:Gdelta}, i.e., $Q \in \mathcal{G}_{S,W}^{\delta}$; and, in~\eqref{rate-distortion-function-generic-distribution-Q}, the Shannon mutual information and the expectation are computed with respect to $Q P_{\hat{W}|S}$. In the case in which $\mathcal{S}\times \mathcal{W}$ is discrete, a result of~\cite[Theorem~10]{Sefidgaran2022} states that every algorithm $\mathcal{A}$ that induces $P_{S,W}$  is $(R(\delta,\epsilon)+\nu_1,\epsilon+\nu_2,\delta;\tilde{d}_m)$-compressible, for every $\nu_1,\nu_2>0$. Combined, the mentioned two results yield the following tail bound on the generalization error for the case of discrete $\mathcal{S}\times \mathcal{W}$,
\begin{equation}
	\gen(S,W)\leq  \sqrt{2\sigma^2(R(\delta,\epsilon)+\log(1/\delta))/n}+\epsilon.
	\label{tail-bound-Sefidgaran-et-al-theorem10}
\end{equation}

It is important to note that the dependence of the tail-bound~\eqref{tail-bound-Sefidgaran-et-al-theorem10} on the input data $S$ is only through the joint distribution $P_{S,W}$, not the particular realization at hand. Because of this, the approach of~\cite{Sefidgaran2022} falls short of accommodating any meaningful connection between their framework and ones that achieve data-dependent bounds such as PAC-Bayes bounds and data-dependent intrinsic dimension-based bounds. In fact, in the terminology of information-theoretic rate-distortion, the described framework can be thought of as being one for fixed-size compressibility, whereas one would here need a framework that allows \textit{variable-size} compressibility. It is precisely such a framework that we develop in this paper. For the ease of the exposition, hereafter we first illustrate our approach and its utility for a simple case. More general results enabled by our approach will be given in the next section. To this end, define 
\begin{align}
	d(w,\hat{w};s)\coloneqq& \gen(s,w)^2-\gen(s,\hat{w})^2, \quad
	d_m(w^m,\hat{w}^m;s^m)\coloneqq \frac{1}{m} \sum \nolimits_{i\in [m]} d(w_i,\hat{w}_i;s_i).
	\label{distortion-measure}
\end{align}

\begin{definition}[Variable-size compressibility] \label{def:varComp} The learning algorithm $\mathcal{A}$ is called  $(R_{S,W}{,}\epsilon{,}\delta;d_m)$-compressible for some $\{R_{s,w}\}_{(s,w)\in \mathcal{S} \times \mathcal{W}}$, where $R_{s,w}\in \mathbb{R}^+$ and $R_{\max}\coloneqq \sup_{s,w} R_{s,w} <\infty$, $\epsilon \in \mathbb{R}$, and $\delta \in \mathbb{R}^+$, if there exists a sequence of hypothesis books $\{\mathcal{H}_m\}_{m\in \mathbb{N}}$, $\mathcal{H}_m {\coloneqq} \{\hat{\vc{w}}[j], j \in [\lfloor e^{m R_{\max}}\rfloor]\}$, such that 
	\begin{align}
		\lim_{m\to \infty} \left[-\frac{1}{m}\log\mathbb{P}_{(S,W)^{\otimes m}}\left(\min_{j \leq e^{\sum_{i\in [m]}R_{S_i,W_i}}} d_m(W^m,\hat{\vc{w}}[j];S^m) >\epsilon \right)\right]\geq \log(1/\delta). \label{eq:condComp}
	\end{align}	
\end{definition}
The former compressibility definition (Definition~\ref{def:compressOld}) corresponds to $R_{s,w}\coloneqq R$ for all $(s,w)$. Comparatively, our Definition~\ref{def:varComp} here accommodates \emph{variable-size} hypothesis books. That is, the number of hypothesis outputs of $\mathcal{H}_m$ among which one searches for a suitable \emph{covering} of $(s^m,w^m)$ depends on $(s^m,w^m)$. The dependency is not only through $P_{S,W}$ but, more importantly, via the quantity $\sum_{i\in [m]}R_{S_i,W_i}$. The theorem that follows, proved in Appendix~\ref{pr:compressibility}, shows how this framework can be used to obtain a data-dependent tail bound on the generalization error. 

\begin{theorem} \label{th:compressibility} If the algorithm $\mathcal{A}$ is $(R_{S,W}{,}\epsilon{,}\delta;d_m)$-compressible and $\forall w{\in} \mathcal{W}$, $\ell(Z,w)$ is $\sigma$-subgaussian, then with probability at least $(1-\delta)$, $$\gen(S,W)\leq \sqrt{4\sigma^2(R_{S,W}+\log(\sqrt{2n}/\delta))/(2n-1)+\epsilon}.$$
\end{theorem}

Note that the seemingly benign generalization to variable-size compressibility has far-reaching consequences for the tail bound itself as well as its proof. For example, notice the difference with the associated bound~\eqref{tail-bound-Sefidgaran-et-al-theorem9} allowed by fixed-size compressibility, especially in terms of the evolution of the bound with the size $n$ of the training dataset. Also, investigating the proof and contrasting it with that of~\eqref{tail-bound-Sefidgaran-et-al-theorem9} for the fixed-size compressibility setting, it is easily seen that while for the latter it is sufficient to consider the union bound over all hypothesis vectors in $\mathcal{H}_m$, among which there exists a suitable \emph{covering} of $(S^m,W^m)$ with probability at least $(1-\delta)$, in our variable-size compressibility case this proof technique does not apply and falls short of producing the desired bound as the \emph{effective} size of the hypothesis book depends on each $(S^m,W^m)$.

Next, we establish a bound on the degree of compressibility of each learning algorithm, proved in Appendix~\ref{pr:covering}.

\begin{theorem} \label{th:covering} Suppose that the algorithm $\mathcal{A}(S)=W$ induces $P_{S,W}$ and $\mathcal{S}\times \mathcal{W}$ is a finite set. Then, for any arbitrary $\nu_1,\nu_2>0$,  $\mathcal{A}$ is $(R_{S,W}+\nu_1,\epsilon+\nu_2,\sigma;d_m)$-compressible if the following sufficient condition holds: for any $\nu_{S,W} \in \mathcal{G}^{\delta}_{S,W}$, 
	\begin{align}
		\inf_{p_{\hat{W}|S}\in \mathcal{Q}(\nu_{S,W})}\inf_{q_{\hat{W}}}  \left\{D_{KL}\left(p_{\hat{W}|S} \nu_{S}\|q_{\hat{W}} \nu_{S}\right)-D_{KL}(\nu_{S,W}\|P_{W|S}\nu_S)\right\}  \leq \mathbb{E}_{\nu_{S,W}}[R_{S,W}],
		\label{eq:ConditionRenyiSimp}
	\end{align}
	where $\mathcal{Q}(\nu_{S,W})$ stands for the set of distributions $p_{\hat{W}|S}$ that satisfy 
	\begin{align}
		\mathbb{E}_{\nu_{S,W} p_{\hat{W}|S}}\left[\gen(S,W)^2-\gen(S,\hat{W})^2\right] \leq \epsilon. \label{eq:tailDistortionSimp}
	\end{align}
\end{theorem}

Combining Theorems~\ref{th:compressibility} and \ref{th:covering} one readily gets a potentially data-dependent tail bound on the generalization error. Because the result is in fact a special case of the more general Theorem~\ref{th:generalTail} that will follow in the next section, we do not state it here. Instead, we elaborate on a useful connection with 
the PAC-Bayes bound of~\cite{mcallester1998some,mcallester1999pac}. For instance, let $P$ be a fixed prior on $\mathcal{W}$. It is not difficult to see that the choice $R_{S,W}\coloneqq  \log \frac{\der P_{W|S}}{\der P}(W)$ satisfies the condition~\eqref{eq:ConditionRenyiSimp} for $\epsilon=0$. The resulting tail bound recovers the PAC-Bayes bound of~\cite{blanchard2007occam,catoni2007pac}, which is a \emph{disintegrated} version of that of~\cite{mcallester1998some,mcallester1999pac}. 

We hasten to mention that an appreciable feature of our approach here, which is rate-distortion theoretic in nature, is its \textit{flexibility}, in the sense that it can be accommodated easily to encompass various forms of tail bounds, by replacing~\eqref{distortion-measure} with a suitable choice of the distortion measure. For example, if instead of a tail bound on the generalization error itself, one seeks a tail bound on the expected generalization error relative to $W \sim \pi$, it suffices to consider $(R_{S,\pi},\epsilon,\delta;d_m)$-compressibility, for some $R_{S,\pi} \in \mathbb{R}^+$, to hold when in the inequality~\eqref{eq:condComp} the left-hand side (LHS) is substituted with
\begin{align*}
	\lim_{m\to \infty} \bigg[{-\frac{1}{m}}\log\mathbb{P}_{S^{\otimes m}}\bigg(\min_{j \leq e^{\sum_{i}R_{S_i,\pi_{S_i}}}} \frac{1}{m}\sum_{i\in [m]} \left(\mathbb{E}_{W_i \sim \pi_{S_i}}[\gen(S_i,W_i)^2]-\gen(S_i,\hat{w}_i[j])^2\right)>\epsilon \bigg)\bigg];
\end{align*}	
and the inequality should hold for any choice of distributions $\pi_S$ (indexed by $S$) over $W$ and any distribution $\nu_{S} \in \mathcal{G}_{S}^{\delta}$ -- Note the change of distortion measure~\eqref{distortion-measure} which now involves an expectation w.r.t. $W \sim \pi_S$. Using this, we obtain that with probability at least $(1-\delta)$, the following holds: 
\begin{equation}
	\forall \pi: \mathbb{E}_{W \sim \pi} [\gen(S,W)] \leq \sqrt{4\sigma^2 (R_{S,\pi}+\log(\sqrt{2n}/\delta)/(2n-1))}.
\end{equation}
A general form of this tail bound which, in particular, recovers as a special case the PAC-Bayes bound of~\cite{mcallester1998some,mcallester1999pac} is stated in Theorem~\ref{th:generalTailExp} and proved in Section~\ref{sec:PAC}.

\section{General data-dependent bounds on the generalization error} \label{sec:generalBounds}
In this section, we take a bigger view. We provide generic bounds, as well as proof techniques to establishing them, that are general enough and apply not only to the generalization error but also to any arbitrary function of the pair $(S,W)$. Specifically, let $f \colon \mathcal{S} \times \mathcal{W} \to \mathbb{R}$ be a given function.  We establish tail bounds, tail bounds on the expectation, and in-expectation bounds (in Appendix~\ref{sec:expBound}) on the random variable $f(S,W)$ that are \textit{in general}   \footnote{However, we hasten to mention that since our approach and the resulting general-purpose bounds of this section are meant to unify several distinct approaches, some of which are data-independent, special instances of our bounds obtained by specialization to those settings can be data-independent.} data-dependent. We insist that by ``data-dependent" we here mean that the bound can be computed using just one sample $S = (Z_1, \ldots, Z_n)$ and does not require knowledge of $P_{S}$. For instance, bounds that depend on $(S,W)$ through its distribution $P_{S,W}$, such as those of~\cite{xu2017information,Sefidgaran2022}, are, in this sense, \textit{data-independent}. Also, as it is shown in Section~\ref{sec:applications}, many existing data-dependent PAC-Bayes and intrinsic dimension-based bounds can be recovered as special cases of our bounds, through judicious choices of $f(S,W)$, e.g., $f(S,W) =  (\gen(S,W))^2$. Moreover, our general-purpose results of this section, which are interesting in their own right, can be used to derive practical novel data-dependent bounds (see Proposition~\ref{prop:PAClossy} and Theorem~\ref{th:RDProcess}).

For the ease of the exposition, our results are stated in terms of a function, denoted as $\mathfrak{T}$, defined as follows. Let $\alpha\geq 1$ and $P_{S}$ and $\nu_{S}$ be two distributions defined over $\mathcal{S}$. Also, let $p_{\hat{W}|S}$ and $q_{\hat{W}|S}$ be two conditional distributions defined over $\hat{\mathcal{W}}$ conditionally given $S$, and $g(S,\hat{W})\colon \mathcal{S} \times \mathcal{\hat{W}} \to \mathbb{R}$ a given function. Then, denote
\begin{align}
	\mathfrak{T}_{\alpha,P_{S}}(\nu_{S},p_{\hat{W}|S},q_{\hat{W}|S},g)\coloneqq & \: \mathbb{E}_{\nu_S}\Big[D_{\alpha}\big(p_{\hat{W}|S}\|q_{\hat{W}|S}\big)\Big]+ \log\mathbb{E}_{P_S q_{\hat{W}|S}}\Big[e^{g(S,\hat{W})}\Big]. \label{def:T}
\end{align}
For $\alpha=1$, as already mentioned, the Rényi divergence coincides with the KL-divergence; and, so, the first term, in that case, is $D_{KL}\big(p_{\hat{W}|S} \nu_{S}\|q_{\hat{W}|S} \nu_{S}\big)$.

\subsection{Tail Bound} \label{sec:generalTail}
Recall the definitions~\eqref{def:Gdelta}, \eqref{def:fSupport}, and~\eqref{def:T}. The following theorem states the main \emph{data-dependent} tail bound of this paper. The bound is general and will be specialized to specific settings in the next section. The proof of the theorem is given in Appendix~\ref{pr:generalTail}.

\begin{theorem} \label{th:generalTail} Let $f(S,W)\colon \mathcal{S} \times \mathcal{W} \to \mathbb{R}$ and $\Delta(S,W)\colon \mathcal{S} \times \mathcal{W} \to \mathbb{R}^+$. Fix arbitrarily the set $\hat{\mathcal{W}}$\footnote{A common choice is to consider $\hat{\mathcal{W}}\subseteq \mathcal{W}$, as in the previous section.} and define arbitrarily $g(S,\hat{W})\colon \mathcal{S} \times \mathcal{\hat{W}} \to \mathbb{R}$. Then, for any $\delta \in \mathbb{R}^+$, with probability at least $1-\delta$,  
	\begin{align}
		f(S,W)\leq \Delta(S,W), \label{eq:tail}
	\end{align}
	if either of the following two conditions holds:
	\begin{itemize}[leftmargin=*]
		\item[i.] For some  $\epsilon \in \mathbb{R}$\footnote{Although for simplicity $\epsilon$ is assumed to take a fixed value here, in general, it can be chosen to depend on $\nu_{S,W}$.}  and any $\nu_{S,W} \in \mathcal{F}_{S,W}^{\delta}\coloneqq \mathcal{G}_{S,W}^{\delta}\bigcap\mathcal{S}_{S,W}\left(f(s,w)- \Delta(s,w)\right)$, it holds that
		\begin{align}
			\inf_{p_{\hat{W}|S}\in \mathcal{Q}(\nu_{S,W})}\inf_{\lambda > 0,q_{\hat{W}|S}}  \Big\{\mathfrak{T}_{1,P_{S}}(\nu_{S},p_{\hat{W}|S},q_{\hat{W}|S}&,\lambda g) {-}D_{KL}(\nu_{S,W} \|P_{W|S}\nu_S) {-} \lambda \left( \mathbb{E}_{\nu_{S,W}}\left[\Delta(S,W)\right]-\epsilon\right)\Big\} \leq \log(\delta),\label{eq:ConditionRenyi}
		\end{align}
		where $\nu_S$ is the marginal distribution of $S$ under  $\nu_{S,W}$	and $\mathcal{Q}(\nu_{S,W})$ is the set of conditionals $p_{\hat{W}|S}$ that satisfy 
		\begin{align}
			\mathbb{E}_{\nu_{S,W} p_{\hat{W}|S}}\big[\Delta(S,W)-g(S,\hat{W})\big] \leq \epsilon. \label{eq:tailDistortion}
		\end{align}
		\item[ii.] For some $\alpha>1$ and any $\nu_{S,W} \in \mathcal{F}_{S,W}^{\delta}$, it holds that
		\begin{align}
			\inf_{q_{W|S},\lambda \geq \alpha/(\alpha-1)}  \Big\{\mathfrak{T}_{\alpha,P_{S}}(\nu_{S},\nu_{W|S},q_{W|S},\lambda \log(f))-&D_{KL}(\nu_{S,W} \|P_{W|S}\nu_S) \nonumber \\
			&-\lambda \left( \mathbb{E}_{\nu_{S}}\log \mathbb{E}_{\nu_{W|S}}\left[\Delta(S,W)\right]\right)\Big\} \leq \log(\delta).
			\label{eq:ConditionRenyiImproved}
		\end{align}
	\end{itemize}
\end{theorem}

It is easily seen that the result of Theorem~\ref{th:generalTail} subsumes that of Theorem~\ref{th:covering}, which can then be seen as a specific case. The compressibility approach that we undertook for the proof of Theorem~\ref{th:covering} can still be used here, with suitable amendments. In particular, the bound of Theorem~\ref{th:generalTail} requires a condition to hold for every $\nu_{s,w}\in \mathcal{F}_{S,W}^{\delta}\subseteq \mathcal{G}_{S,W}^{\delta}$. This is equivalent to \emph{covering} all sequences $(S^m,W^m)$ whose empirical distributions $Q$ are in the vicinity of $P_{S,W}$ in the sense of~\eqref{def:Gdelta}, using the $\hat{W}$ defined by $p_{\hat{W}|S}$. Furthermore, the distribution $q_{\hat{W}|S}$ is the one used to build (part of) the hypothesis book $\mathcal{H}_{m,Q}$ (see the proof of Theorem~\ref{th:covering} for a definition of $\mathcal{H}_{m,Q}$).

The bound \eqref{eq:tail} holds with probability $(1-\delta)$ with respect to the joint distribution $P_{S,W}$. It should be noted that if a learning algorithm $\mathcal{A}(S)$ induces $P_{S,W}$, one can consider the bound~\eqref{eq:tail} with respect to any alternate distribution $\tilde{P}_{S,W}=P_S\tilde{P}_{W|S}$ and substitute accordingly in the conditions of the theorem, as is common in PAC-Bayes literature, see, e.g., \cite{dziugaite2017computing,neyshabur2018pacbayesian}.

Furthermore, in our framework of compressibility, $\epsilon$ stands for the allowed level of average distortion in~\eqref{eq:tailDistortion}. The specific case $\epsilon=0$ corresponds to lossless compression; and, it is clear that allowing a non-zero average distortion level, i.e., $\epsilon \neq 0$, can yield a tighter bound~\eqref{eq:tail}. In fact, as will be shown in the subsequent sections, many known data-dependent PAC-Bayes and intrinsic dimension-based bounds can be recovered from the ``lossless compression'' case. Also, for $\epsilon=0$ the condition~\eqref{eq:ConditionRenyi} of the part (i.) of the theorem with the choices $\mathcal{\hat{W}}\coloneqq\mathcal{W}$, $g(s,\hat{w})\coloneqq f(s,\hat{w})$, and $p_{\hat{W}|S}\coloneqq \nu_{W|S}$ reduces to
\begin{align}
	\inf_{q_{W|S},\lambda > 0}  \bigg\{& \mathbb{E}_{\nu_{S,W}}\left[ \log\left(\frac{\der P_{W|S}}{\der q_{W|S}}\right){+} \log\mathbb{E}_{P_S q_{W|S}}\left[e^{\lambda f(S,W)}\right]{-}\lambda \Delta(S,W)\right]\bigg\} \leq \log(\delta).
	\label{eq:ConditionRenyiSimpLoss}
\end{align}
Besides, since $\mathcal{F}_{S,W}^{\delta}\subseteq \mathcal{G}_{S,W}^{\delta}$, the theorem holds if we consider all $\nu_{S,W} \in \mathcal{G}_{S,W}^{\delta}$. The latter set, although possibly larger, seems to be more suitable for analytical investigations. Furthermore, for any $\nu_{S,W} \in  \mathcal{F}_{S,W}^{\delta}$, the distortion criterion \eqref{eq:tailDistortion} is satisfied whenever 
\begin{align}
	\mathbb{E}_{\nu_{S,W} p_{\hat{W}|S}}\big[f(S,W)-g(S,\hat{W})\big] \leq \epsilon. \label{eq:tailDistortion2}
\end{align}
This condition is often easier to consider, as we will see in the next section. In particular, \eqref{eq:tailDistortion2} can be further simplified under the Lipschitz assumption, \ie when $\forall w, \hat{w}, s \colon  |f(s,w)-g(s,\hat{w})| \leq \mathfrak{L} \rho(w,\hat{w})$, where $\rho\colon \mathcal{W} \times \hat{\mathcal{W}} \to \mathbb{R}^+$ is a distortion measure over $\mathcal{W} \times \hat{\mathcal{W}}$. In this case, a sufficient condition to meet the distortion criterion \eqref{eq:tailDistortion} is 
\begin{align}
	\mathbb{E}_{\nu_{S,W} p_{\hat{W}|S}}\big[\rho(W,\hat{W})\big] \leq  \epsilon/(2\mathfrak{L}). 
	\label{eq:tailDistortion3}
\end{align}
We close this section by mentioning that the result of Theorem~\ref{th:generalTail} can be extended easily to accommodate any additional possible stochasticity of the algorithm, i.e., such as in~\cite{harutyunyan2021}. The result is stated in Corollary~\ref{cor:generalTailU} in the appendices. 


\subsection{Tail bound on the expectation} \label{sec:tailExp}

In this section, we establish a potentially data-dependent tail bound on the expectation of $f(S,W)$. That is, the bound is on $\mathbb{E}_{W\sim \pi}\left[f(S,W)\right]$ for every distribution $\pi$ over $\mathcal{W}$.\footnote{The result trivially holds if the choices of $\pi$ are restricted to a given subset of distributions, as used in Theorem~\ref{th:RDProcess}.} 

\begin{theorem} \label{th:generalTailExp} Let $f(S,W)\colon \mathcal{S} \times \mathcal{W} \to \mathbb{R}$ and $\Delta(S,\pi)\colon \mathcal{S} \times \mathcal{P}_{\mathcal{W}} \to \mathbb{R}^+$. Fix any set $\hat{\mathcal{W}}$ and define arbitrarily $g(S,\hat{W})\colon \mathcal{S} \times \mathcal{\hat{W}} \to \mathbb{R}$.  Then, for any $\delta \in \mathbb{R}^+$, with probability at least $1-\delta$,  
	\begin{align}
		\forall \pi: \mathbb{E}_{W\sim \pi}\left[f(S,W)\right]\leq \Delta(S,\pi),
	\end{align}
	if either of the following two conditions holds: 
	\begin{itemize}[leftmargin=*]
		\item[i.] For $\epsilon \in \mathbb{R}$ and for any choice of distributions $\pi_S$ (indexed by $S$) over $\mathcal{W}$ and any distribution $\nu_{S} \in \mathcal{F}_{S,\pi_S}^{\delta} \coloneqq $ \\$ \mathcal{G}_{S}^{\delta}\bigcap\mathcal{S}_{S}\left( \mathbb{E}_{W \sim \pi_s}[f(s,W)]- \Delta(s,\pi_s)\right)$,
		\begin{align}
			\inf_{p_{\hat{W}|S}\in \mathcal{Q}( \nu_{S} \pi_S)}\inf_{q_{\hat{W}|S},\lambda > 0}  \Big\{&		\mathfrak{T}_{1,P_S}(\nu_S,p_{\hat{W}|S},q_{\hat{W}|S},\lambda g)-\lambda \left( \mathbb{E}_{\nu_{S}}\left[\Delta(S,\pi_S)\right]-\epsilon\right) \Big\} \leq \log(\delta), \label{eq:ConditionTailExp}
		\end{align}
		where  $\mathfrak{T}_{\alpha,P_S}$ is defined in \eqref{def:T} and $\mathcal{Q}(\nu_{S}\pi_S)$ contains all the distributions $p_{\hat{W}|S}$ such that 
		\begin{align}
			\mathbb{E}_{\nu_{S} p_{\hat{W}|S}}\big[\Delta(S,\pi_S)-g(S,\hat{W})\big] \leq \epsilon. \label{eq:distExp}
		\end{align}  
		\item[ii.] For some $\alpha>1$ and any choice of distributions $\pi_S$ over $\mathcal{W}$ and any distribution $\nu_{S} \in \mathcal{F}_{S,\pi_S}^{\delta}$,
		\begin{align}
			\inf_{q_{W|S},\lambda \geq \alpha/(\alpha-1)}  \Big\{&\mathfrak{T}_{\alpha,P_S}(\nu_{S},\pi_S,q_{W|S},\lambda\log(f)) -\lambda \mathbb{E}_{\nu_{S}}\left[\log \Delta(S,\pi_S)\right]\Big\} \leq \log(\delta).\label{eq:ConRenTailExpImproved}
		\end{align}
	\end{itemize}	
\end{theorem} 
The theorem is proved in Appendix~\ref{pr:generalTailExp}. The result, which is interesting in its own right, in particular, allows us to establish a formal connection between the variable-size compressibility approach that we develop in this paper and the seemingly unrelated PAC-Bayes approaches. In fact, as we will see in Section~\ref{sec:PAC}, several known PAC-Bayes bounds can be recovered as special cases of our Theorem~\ref{th:generalTailExp}. It should be noted that prior to this work such connections were established only for a few limited settings, such as the compressibility framework of \cite{littlestone1986relating}, which differs from ours, and for which connections with PAC-Bayes approaches were established by Blum and Langford \cite{blum2003pac}.

Furthermore, observe that for $\epsilon=0$, the condition \eqref{eq:ConditionTailExp} with the choices $\mathcal{\hat{W}}\coloneqq\mathcal{W}$, $g(s,\hat{w})\coloneqq f(s,w)$, and $p_{\hat{W}|S}\coloneqq \nu_{W|S}$ yields
\begin{align}
	\inf_{q_{W|S},\lambda > 0}  \bigg\{& D_{KL}\left(\pi_S \nu_{S}\|q_{W|S} \nu_{S}\right)+ \log\mathbb{E}_{P_S q_{W|S}}\left[e^{\lambda f(S,W)}\right]-\lambda  \mathbb{E}_{\nu_{S}}\left[\Delta(S,\pi_S)\right]\bigg\} \leq \log(\delta). \label{eq:ConditionTailExpSimp}
\end{align}


\subsection{In-expectation bound}  \label{sec:expBound}
Finally, here, we propose our bound on the expectation of $f(S,W)$.

\begin{theorem} \label{th:expectation} Let $f(S,W)\colon \mathcal{S} \times \mathcal{W} \to \mathbb{R}$. Fix arbitrarily the set $\hat{\mathcal{W}}$ and define arbitrarily $g(S,\hat{W})\colon \mathcal{S} \times \mathcal{\hat{W}} \to \mathbb{R}$. 	
	\begin{itemize}[leftmargin=*]
		\item[i.] For any $\epsilon \in \mathbb{R}$,
		\begin{align*}
			\mathbb{E}_{(S,W)\sim P_{S,W}}\left[f(S,W)\right]\leq 
			\inf_{p_{\hat{W}|S}\in \mathcal{Q}(P_{S,W})}\inf_{q_{\hat{W}|S},\lambda > 0}  \big\{\frac{1}{\lambda}\mathfrak{T}_{1,P_S}(P_S,p_{\hat{W}|S},q_{\hat{W}|S},\lambda g) \big\} +\epsilon,
		\end{align*}
		where  $\mathfrak{T}_{1,P_S}$ is defined in \eqref{def:T} and $\mathcal{Q}(P_{S,W})$ contains all the distributions $p_{\hat{W}|S}$ that satisfy $\mathbb{E}_{P_{S,W} p_{\hat{W}|S}}[f(S,W)-g(S,\hat{W})] \leq \epsilon$.
		\item[ii.] For $\alpha >1$, 
		\begin{align*}
		\hspace{-0.2 cm}	\hspace{-0.1 cm}\log \mathbb{E}_{(S,W)\sim P_{S,W}}\left[f(S,W)\right]\leq 
			\inf_{q_{W|S},\lambda \geq \alpha/(\alpha-1)}  \Big\{\frac{1}{\lambda}D_{\alpha} \left(P_{S,W}\|q_{W|S}P_S\right)+ \frac{1}{\lambda} \log\mathbb{E}_{P_Sq_{W|S}}\left[f(S,W)^{\lambda }\right] \Big\}.
		\end{align*}
	\end{itemize} 
	
\end{theorem}
The theorem is proved in Appendix~\ref{pr:expectation}. It can be easily checked that part i. of this result includes the mutual information-based results. For example, one easily recovers the result of~\cite[Theorem~1]{xu2017information} by setting $\hat{W}=W$,  $q_{\hat{W}|S}$ as the marginal distribution of $W$ under $P_{S,W}$,  $\lambda=\sqrt{\frac{2nI(S;W)}{\sigma^2}}$ and $f(S,W)=g(S,W)=\gen(S,W)$ or $f(S,W)=g(S,W)=-\gen(S,W)$. This can be also extended to recover its single-datum counterpart \cite{Bu2020}. In Appendix~\ref{sec:limitations} we also show that the result of Theorem~\ref{th:expectation} escapes the issues mentioned in~\cite{haghifam2023limitations} as limitations of information-theoretic generalization bounds.

 
\section{Applications} \label{sec:applications}
In this section, we apply our general bounds of Section~\ref{sec:generalBounds} to various settings; and we show how one can recover, and sometimes obtain \textit{new and possibly tighter}, \textit{data-dependent} bounds. The settings that we consider include rate-distortion theoretic, PAC-Bayes, and dimension-based approaches. As these have so far been thought of, and developed, largely independently of each other in the related literature, in particular, this unveils the strength and unifying character of our variable-size compression framework. We remind the reader that the fixed-size compressibility approach of~\cite{Sefidgaran2022}, which only allows obtaining \textit{data-independent} bounds, is not applicable here.


\subsection{Rate-distortion theoretic bounds}
We show how the rate-distortion theoretic tail bound of \cite[Theorem~10]{Sefidgaran2022} can be recovered using our Theorem~\ref{th:generalTail}. Let $f(S,W)\coloneqq \gen(S,W)$, $\mathcal{\hat{W}}\subseteq \mathcal{W}$, and  \begin{align}
    \Delta(S,W)\coloneqq \Delta \coloneqq \sqrt{2\sigma^2\Big(\sup_{\nu_{S,W}\in \mathcal{G}_{S,W}^{\delta}}\mathfrak{RD}(\epsilon;\nu_{S,W})+\log(1/\delta)\Big)/n}+\epsilon. \label{eq:RDBound}
\end{align} 
Now, let $g(s,\hat{w})\coloneqq \gen(s,\hat{w})$, $\epsilon \in \mathbb{R}$, $\alpha=1$, and $q_{\hat{W}|S}\coloneqq q_{\hat{W}}$. Then, for any $\nu_{S,W} \in \mathcal{G}_{S,W}^{\delta}$, 
	\begin{align*}
		\inf_{p_{\hat{W}|S}\in \mathcal{Q}(\nu_{S,W})}\inf_{q_{\hat{W}},\lambda \geq 0}  \Big\{D_{KL}\Big(p_{\hat{W}|S} \nu_S\|&q_{\hat{W}} \nu_S\Big) -\lambda \left( \mathbb{E}_{\nu_{S,W}}\left[\Delta\right]-\epsilon\right)+\log\mathbb{E}_{P_S q_{\hat{W}}}\left[e^{\lambda g(S,\hat{W})}\right]\Big\}\\ 
		&\leq \inf_{p_{\hat{W}|S}\in \mathcal{Q}(\nu_{S,W})}\inf_{\lambda \geq 0}  \Big\{I(S;\hat{W}) -\lambda \left( \mathbb{E}_{\nu_{S,W}}\left[\Delta\right]-\epsilon\right)+\frac{
			\lambda^2 \sigma^2}{2n}\Big\}\\ 
		&\leq 	\inf_{\lambda \geq 0}  \left\{\mathfrak{RD}(\epsilon;\nu_{S,W})-\lambda \left( \Delta-\epsilon\right)+\frac{
			\lambda^2 \sigma^2}{2n}\right\},
	\end{align*}
	which is less than $\log(\delta)$ by letting $\lambda \coloneqq n(\Delta-\epsilon)/\sigma^2$. Hence, since $D_{KL}(\nu_{S,W}\|P_{W|S}\nu_S)\geq 0$, part i. of Theorem~\ref{th:generalTail} yields \cite[Theorem~10]{Sefidgaran2022} with the average distortion criterion. 

Furthermore, Theorem~\ref{th:generalTail} allows to extend this result in various ways. An example is establishing bounds for any $\alpha \geq 1$. As another example, consider $\epsilon=0$ and $f(s,w)\coloneqq n D_{KL}(\mathcal{\hat{L}}(w)\|\mathcal{L}(s,w))$, where for $a,b>0$, $D_{KL}(a\|b)\coloneqq a\log(a/b)+(1-a)\log((1-a)/(1-b))$. This choice was used previously by \cite{seeger2002pac} to derive PAC-Bayes bound. Now, using the following inequality of \cite{tolstikhin2013pac}, $D_{KL}^{-1}(a\|b) \leq a +\sqrt{2ab}+2b$, where $D_{KL}^{-1}(a\|b)\coloneqq \sup \left\{p\in [0,1]\colon D_{KL}(p\|a) \leq b \right\}$, we establish that, with probability $1-\delta$, $	\gen(S,W) \leq \sqrt{\mathcal{\hat{L}}(S,W)C/n}+C/n$, where $C \coloneqq 4\sigma^2\big(\sup \nolimits_{\nu_{S,W}\in \mathcal{G}_{S,W}^{\delta}}I(S;W)+\log(2\sqrt{n}/\delta)\big)$. This bound achieves the generalization bound with $\mathcal{O}(1/n)$, if $\mathcal{\hat{L}}(S,W)=0$. A similar approach can be taken for the lossy case, as used in the PAC-Bayes approach by Biggs and
Guedj \cite{biggs2022margins}.


\subsection{PAC-Bayes bounds} \label{sec:PAC}

PAC-Bayes bounds were introduced initially by McAllester~\cite{mcallester1998some,mcallester1999pac}; and, since then, developed further in many other works. The reader is referred to \cite{guedj2019primer,alquier2021} for a summary of recent developments on this. In this section, we show that our general framework not only recovers several existing PAC-Bayes bounds (in doing so, we focus on the most general ones) but also allows to derive novel ones.

Consider part i.~of Theorem~\ref{th:generalTailExp} and the condition~\eqref{eq:ConditionTailExpSimp}. Let $\lambda \coloneqq 1$ and set $q_{W|S}$ to be a fixed, possibly data-dependent, common to all $\nu_{S,W}$. This yields that with probability at least $(1-\delta)$ we have
\begin{align}
	\forall \pi\colon \mathbb{E}_{W\sim \pi}\left[f(S,W)\right] \leq  D_{KL}\left(\pi \|q_{W|S} \right)+ \log\mathbb{E}_{P_S q_{W|S}}  \left[e^{ f(S,W)}\right]+\log{(1/\delta)}. \label{eq:ConditionTailExpSimp2}
\end{align}
The obtained bound~\eqref{eq:ConditionTailExpSimp2} equals that of~\cite[Theorem~1.ii]{rivasplata2020pac}. Similarly, derivations using our Theorem~\ref{th:generalTail} and the condition~\eqref{eq:ConditionRenyiSimpLoss} allow to recover the result of~\cite[Theorem~1.i]{rivasplata2020pac}. As observed by Clerico et al.~\cite{clerico2022pac}, these recovered bounds are themselves general enough to subsume most of other existing PAC-Bayes bounds including those of~\cite{mcallester1998some,mcallester1999pac,seeger2002pac,catoni2003pac,maurer2004note,catoni2007pac,germain2009pac,tolstikhin2013pac,thiemann2017strongly}. Similarly, part ii. of Theorem~\ref{th:generalTailExp} with the choice $\lambda = \alpha/(\alpha-1)$ allows to recover the result of~\cite{begin2016pac} (see also \cite[Theorem~1]{viallard2021general}). 

Our variable-size compressibility framework and our general bounds of Section~\eqref{sec:generalBounds} also allow to establish \textit{novel} PAC-Bayes type bounds. The following proposition, proved in Appendix~\ref{pr:PAClossy}, provides an example of such bounds. 

\begin{proposition} \label{prop:PAClossy} Let the set $\mathcal{\hat{W}}$ and function $g(S,\hat{W})$ be arbitrary.  Let $q_{\hat{W}|S}$ be a possibly data-dependent prior over $\hat{\mathcal{W}}$.
	\begin{itemize}[leftmargin=*]
		\item[i.]  With probability at least $(1-\delta)$, we have
		\begin{align}
			\forall \pi \colon \mathbb{E}_{\pi}[f(S,W)] \leq  D_{KL}\left(p_{\hat{W}|S,\pi} \|q_{\hat{W}|S} \right)+ \log\mathbb{E}_{P_S q_{\hat{W}|S}}  \left[e^{ g(S,\hat{W})}\right]+\log{(1/\delta)}+\epsilon, \label{eq:ConditionTailExpLoss2}
		\end{align}
		where for any $\pi$, $p_{\hat{W}|S,\pi}$ is such that $\mathbb{E}_{\pi P_{\hat{W}|S,\pi}}\big[f(S,W)-g(S,\hat{W})\big] \leq \epsilon$.
		\item[ii.] Let $p_{\hat{W}|S,W}=p_{\hat{W}|W}$ be any distribution such that for any $\nu_{s,w} \in \mathcal{F}_{S,W}^{\delta}$, $\mathbb{E}_{\nu_{S,W}p_{\hat{W}|W}}\big[\Delta(S,W)-g(S,\hat{W})\big] \leq \epsilon$. Denote the induced conditional distribution of $\hat{W}$ given $S$, under $P_{W|S}p_{\hat{W}|W,S}$, where $p_{\hat{W}|W,S}=p_{\hat{W}|W}$,  as $p^*_{\hat{W}|S}$. 
		Then, with probability at least $1-\delta$,
		\begin{align}
			f(S,W) \leq  \mathbb{E}_{p_{\hat{W}|W}}\bigg[\log\bigg(\frac{\der p^*_{\hat{W}|S}} {\der q_{\hat{W}|S}}\bigg)\bigg] + \log\mathbb{E}_{P_S q_{\hat{W}|S}}  \left[e^{ g(S,\hat{W})}\right]+\log{(1/\delta)}+\epsilon. \label{eq:ConditionTailExpLoss}
		\end{align}
	\end{itemize}
\end{proposition}

 We remark that, unlike the classical PAC-Bayes bounds  of~\cite{mcallester1998some,mcallester1999pac}, our bound of Proposition~\ref{prop:PAClossy} does not become vacuous ($\infty$) for
deterministic algorithms with continuous parameters. To show this, consider a \emph{toy} example that is similar to the one of~\cite[p.11]{Sefidgaran2022}: assume $Z \in \mathbb{R}^d$ and $W=\frac{1}{n}\sum_{i\in[n]}Z_i$. Further, assume that $\ell(Z,w)$ is $\sigma$-subgaussian for every $w\in \mathbb{R}^d$ and $|\ell(Z,w)-\ell(Z,\hat{w})|\leq \mathfrak{L}\|w-\hat{w}\|^2$ for some $\mathfrak{L}\in \mathbb{R}^+$. For this example, it is easy to see that the bounds of \cite{mcallester1998some,mcallester1999pac} take infinite values, whereas our Proposition~\ref{prop:PAClossy} leads to the following upper bound 
\begin{align*}
    \sqrt{2\sigma^2 \left(2\sqrt{\mathfrak{L}d\sum_{j\in[d]}\big(\sum_{i\in [n]} Z_{i,j}/n\big)^2}+\log(1/\delta)\right) \, \Big/ n}.
\end{align*}
This is easily seen by letting $\mathcal{\hat{W}}\subseteq \mathcal{W}$, $\hat{W}=W+N$, where $N\sim\mathcal{N} (0,\epsilon/(2\mathfrak{L}d)\mathrm{I}_d)$, $p_{\hat{W}|S}=\mathcal{N}(W,\epsilon/(2\mathfrak{L}d)\mathrm{I}_d)$, $q_{\hat{W}|S}=\mathcal{N}(0,\epsilon/(2\mathfrak{L}d)\mathrm{I}_d)$ and $f(S,W)=g(S,W)=\lambda \gen(S,W)$, where $\mathrm{I}_d$ is the identity matrix of size $d$.

The bound~\eqref{eq:ConditionTailExpLoss2} is also closely related to results obtained in \cite{langford2001bounds,langford2002pac, biggs2022margins}. In particular, considering $f(S,W)$ as a function of the $\gamma$-margin generalization error and $g(S,\hat{W})$ as a function of the generalization error of the 0-1 loss function with threshold $\gamma/2$, sufficient conditions to meet the distortion criterion are provided in terms of the margin in~\cite{biggs2022margins}. The results are then applied to various setups ranging from SVM to ReLU neural networks. It is noteworthy that, under the Lipschitz assumption for $f(S,W)$, an alternate sufficient condition can be derived. For example, suppose that $\mathcal{\hat{W}} \subseteq \mathcal{W}$ and $f(s,w)=g(s,w)\coloneqq \lambda \gen(s,w)^2$ for some $\lambda >0$. Then for the $B$-bounded and $\mathfrak{L}$-Lipschitz loss function, \ie if $\forall z\in \mathcal{Z}, w,w' \in \mathcal{W}\colon |\ell(z,w)-\ell(z,w')|\leq \mathfrak{L} \rho(w,w')$, where $\rho\colon \mathcal{W}\times \mathcal{W}\to \mathbb{R}^+$ is a distortion function on $\mathcal{W}$, a sufficient condition for satisfying the distortion criterion is that $\mathbb{E}_{\nu_{W}P_{\hat{W}|S}}[\rho(W,\hat{W})] \leq 4B\mathfrak{L}\lambda \epsilon$ holds true for every $\nu_{s,w} \in \mathcal{G}_{S,W}^{\delta}$. If $\rho(w,\hat{w})\coloneqq\|w-\hat{w}\|$, an example is  $\hat{W}\coloneqq W+Z$, where $Z$ is an independent noise such that $\mathbb{E}\left[\|Z\|\right]\leq 4B\mathfrak{L}\epsilon$. In particular, this choice prevents the upper bound from taking very large (infinite) values.

To the best knowledge of the authors, the bound~\eqref{eq:ConditionTailExpLoss} has \textit{not} been reported in the PAC-Bayes literature; and it appears to be a \textit{novel} one. This bound subsumes that of~\cite[Theorem~1.i]{rivasplata2020pac} which, as can be seen easily, it recovers with the specific choices $\hat{\mathcal{W}}\coloneqq\mathcal{W}$, $g(s,\hat{w})\coloneqq f(s,\hat{w})$ and $p_{\hat{w}|w}\coloneqq \delta_{w}$ (a Dirac delta function). Furthermore, we show in Appendix~\ref{sec:limitations} that this bound can resolve the limitations of the classical PAC-Bayes bounds raised in~\cite{haghifam2023limitations}.


\subsection{Dimension-based bounds}
Prior to this work, the connection between compressibility and intrinsic dimension-based approaches has been established in \cite{Sefidgaran2022}. However, as the framework introduced therein is of a ``fixed-size'' compressibility type and only allows establishing data-independent bounds, the connection was made only to the intrinsic dimensions of the \emph{marginal} distributions introduced by the algorithm. This departs from most of the proposed dimension-based bounds in the related literature, which are data-dependent, i.e., they depend on a particular dimension arising for a given $S=s$. See, e.g., \cite{simsekli2020hausdorff,camuto2021fractal,hodgkinson2022generalization}. 

In Appendix~\ref{sec:HausdorffSGD}, we show how using our variable-size compressibility one can recover the main generalization error bound of \cite{simsekli2020hausdorff}. The approach can be extended similarly to derive \cite[Theorem~1]{camuto2021fractal}.

In what follows, we make a new connection between the generalization error and the rate-distortion of a process arising from the optimization trajectories. Denote the hypothesis along the trajectory at iteration or time $t$ by $\mathcal{A}(S,t)\coloneqq W_S^{t}\coloneqq W^{t}$. The time $t$ can be discrete or continuous time. For continuous-time trajectory, we will use $\T\coloneqq [0,1] \subseteq \mathbb{R}$; and, for discrete time, we use $\T\coloneqq [t_1:t_2]\subseteq \mathbb{N}$. We denote the set of hypotheses along the trajectory by $W^{\T}$. 

{\c S}im{\c s}ekli et al.~\cite{simsekli2020hausdorff} made a connection between the Hausdorff dimension of the  \emph{continuous-time} SGD and $\sup_{t\in [0,1]} \gen(S,W^t_S)$. This finding has been also verified numerically \cite{simsekli2020hausdorff}. However, it is not clear enough at least to us (i) how the supremum over time of the generalization error of the continuous-time model relates to the maximum of the generalization error of the discrete-time model (for which the Hausdorff dimension equals zero), and (ii) how the maximum of the generalization error over the optimization trajectory relates, order-wise, to the generalization error of the final model. For this reason, instead, hereafter we consider a discrete-time model for the optimization trajectory (compatible with common optimization algorithms such as SGD), together with the average of the generalization error along the trajectory. Specifically, let
\begin{align}
	\gen(S,W^{\T}_S)\coloneqq \frac{1}{\dT}\sum\nolimits_{t\in \T} \gen(S,W_S^t),
	\label{objective-function-intrinsic-dimension-optimization-trajectories}
\end{align}
where $\dT \coloneqq t_2 -t_1$, $t_1,t_2 \in \mathbb{N}$. In particular, the quantity~\eqref{objective-function-intrinsic-dimension-optimization-trajectories} hinges on that there is information residing in the optimization trajectories, an aspect that was already observed numerically in \cite{jastrzkebski2017three,jastrzebski2020break,jastrzebski2021catastrophic,martin2021implicit,xing2018walk,hodgkinson2022generalization}. For a given dataset if $t_1$ and $\Delta t_2$ are large, the intuition suggests that \eqref{objective-function-intrinsic-dimension-optimization-trajectories} is close to the ``average'' generalization error of the algorithm. In particular, this holds, if we assume an ergodic invariant measure for the distribution of $W^{\infty}$, as in \cite{camuto2021fractal}.

The results of this section are expressed in terms of the rate-distortion functions of the process. Assume $\mathcal{\hat{W}} \subseteq \mathcal{W}$. For any distribution $\nu_S$ defined over $\mathcal{S}$ and any conditional distribution $Q_{W^{\T}|S}$ of $W^{\T}$ given $S$, define the rate-distortion of the optimization trajectory $\T\coloneqq [t_1:t_2]$ as
\begin{align}
	\mathfrak{RD}(\nu_S,\epsilon;Q_{W^{\T}|S}) \coloneqq  \inf\nolimits _{P_{\hat{W}^{\T}|S}} I(S;\hat{W}^{\T}), \quad \text{s.t.}\quad  \mathbb{E}\big[\gen(S,W^{\T})-\gen(S,\hat{W}^{\T})\big] \leq \epsilon, \label{def:RDProcess1}
\end{align}
where the mutual information and expectation are with respect to $\nu_S Q_{W^{\T}|S}P_{\hat{W}^{\T}|S}$.

Now, we establish a bound on the conditional expectation of $\gen(S,W^{\T})$, which is stated for simplicity for the case $\ell(z,w)\in [0,1]$. The theorem is proved in Appendix~\ref{pr:RDProcessSup}.
\begin{theorem} \label{th:RDProcessSup} Suppose that $\ell\colon \mathcal{Z}\times \mathcal{W}\to [0,1]$. Then, for any $\epsilon \in \mathbb{R}$ with probability at least $1-\delta$, 
	\begin{align*}
		\forall \pi\colon \mathbb{E}_{W^{\T}\sim \pi}\left[\gen(S,W^{\T})\right] \leq & \sqrt{\Big(\sup_{\nu_s \in \mathcal{G}^{\delta}_S}\mathfrak{RD}(\nu_S,\epsilon;\pi)+\log(1/\delta)\Big)/(2n)}+\epsilon.
	\end{align*}
\end{theorem}
This result suggests that in order to have a small average generalization error, the trajectory should be `compressible' for all empirical distributions that are within $\delta$-vicinity of the true unknown distribution $\mu^{\otimes n}$. This bound is not data-dependent, however. Theorem~\ref{th:RDProcess} that will follow provides one that is \textit{data-dependent}; it relates the average generalization error to the rate-distortion of the optimization trajectory, induced for a given $s$.  Assume that the loss is $\mathfrak{L}$-Lipschitz, \ie  $\forall z\in \mathcal{Z}, w,\hat{w} \in \mathcal{W}\colon |\ell(z,w)-\ell(z,\hat{w})|\leq \mathfrak{L} \rho(w,\hat{w})$, where $\rho\colon \mathcal{W}\times \mathcal{\hat{W}}\to \mathbb{R}^+$ is a distortion function on $\mathcal{W}$. Define the rate-distortion of the optimization trajectory $\T\coloneqq [t_1:t_2]$ for any $S=s$ and $P_{W^{\T}|s}$ as
\begin{align}
	\mathfrak{RD}(s,\epsilon;P_{W^{\T}|s}) \coloneqq \inf\nolimits_{P_{\hat{W}^{\T}|W^{\T},s}} I(W^{\T};\hat{W}^{\T}), \quad \text{s.t.}\quad \mathbb{E}\big[\rho(W^{\T},\hat{W}^{\T})\big] \leq \epsilon, \label{def:RDProcess2}
\end{align}
where the mutual information and expectation are with respect to $P_{W^{\T}|s} P_{\hat{W}^{\T}|W^{\T},s}$, and with slight abuse of notations $\rho(W^{\T},\hat{W}^{\T}) \\\coloneqq \frac{1}{\dT} \sum_{t\in \T}\rho(w^t,\hat{w}^t)$ is the average distortion along the iterations. Note that the defined rate-distortion function depends on the distortion function $\rho$, which is assumed to be fixed.

Conceptually, there is a difference between the two rate-distortion functions defined above. Namely, while \eqref{def:RDProcess1} considers the joint \emph{compression} of all trajectories $W^{\T}\sim Q_{W^{\T}|S}$ when $S\sim \nu_S$ \eqref{def:RDProcess2} considers joint \emph{compression} of all iterations $W^{\T}\sim Q_{W^{\T}|s}$ for a particular dataset $s$. Now, we are ready to state the main result of this section. The proof is given in Appendix~\ref{pr:RDProcess}. 
\begin{theorem} \label{th:RDProcess} Suppose that the loss $\ell\colon \mathcal{Z}\times \mathcal{W}\to [0,1]$ is $\mathfrak{L}$-Lipschitz with respect to $\rho\colon\mathcal{W} \times \mathcal{\hat{W}}\to \mathbb{R}^+$. Suppose that the optimization algorithm induces $P_{W^{\T}|S}$, denoted as $\pi_S$ for brevity.
	Then, for any $\epsilon \in \mathbb{R}$ with probability at least $1-\delta$, 
	\begin{align*}
		\mathbb{E}_{W^{\T}\sim \pi_S}\left[\gen(S,W^{\T})\right] \leq &  \sqrt{\big(\mathfrak{RD}(S,\epsilon;\pi_S)+\log(\sqrt{2n} M/\delta)\big)/(2n-1)+4\mathfrak{L}\epsilon},
	\end{align*}
	where $	M \coloneqq \exp \left(\sup\nolimits_{\nu_S \in \mathcal{G}_{S}^{\delta}} \mathbb{E}_{\pi_S}  \log \frac{\der\pi_S}{\der [\pi_S \nu_S]_{W^{\T}}}\right)$ and $[\pi_S \nu_S]_{W^{\T}}$ denotes the marginal distribution of $W^{\T}$ under $\pi_S \nu_S$. 
\end{theorem}
The \emph{coupling} coefficient $M$ is similar to one defined in \cite[Definition~5]{simsekli2020hausdorff} (See Assumption~\ref{a:M} in the Appendix); and intuitively, it measures the dependency of ${W}^T$ on $S$. This term can be bounded by $I_{\infty}(S;W^{\T})$, a quantity that is often considered in related literature \cite{hodgkinson2022generalization,lim2022chaotic}. 

The result of Theorem~\ref{th:RDProcess} establishes a connection between the generalization error and new data-dependent intrinsic dimensions. To see this, let $	R_{\dT}(s,\epsilon;P_{W^{\T}|S})\coloneqq \mathfrak{RD}(s,\epsilon;P_{W^{\T}|S})/\dT$. Assume that the conditional $P_{W^{\T}|S}$ is stationary and ergodic; and consider the following $R_{\infty}(s,\epsilon)\coloneqq \lim_{\dT \ra \infty} 	R_{\dT}(s,\epsilon;P_{W^{\T}|S})$. It has been shown that for the i.i.d. process and also the Gauss-Markov process, the convergence to the infinite limit is with the rate of $1/\sqrt{\dT}$ \cite{kostina2012fixed,tian2019dispersion}. Hence, this limit gives a reasonable estimate of  $R_{\dT}(s,\epsilon;P_{W^{\T}|S})$, even for modest $\dT$.

Rezagah et al.~\cite{Rezagah2016} and Geiger and Koch~\cite{geiger2019information} studied the ratio $\lim_{\epsilon \to 0} \frac{R_{\infty}(s,\epsilon)}{\log(1/\epsilon)}$, known as \emph{rate-distortion dimension of the process}. Using a similar approach as in \cite[Proof of Corollary~7]{Sefidgaran2022}, the generalization bound of Theorem~\ref{th:RDProcessSup} can be expressed in terms of the rate-distortion dimension of the optimization trajectories. Rezagah et al.~\cite{Rezagah2016} and Geiger and Koch~\cite{geiger2019information} have shown that for a large family of distortions $\rho(w,\hat{w})$, the rate-distortion dimension of the process coincides with the Rényi information dimension of the process, and determines the fundamental limits of compressed sensing settings \cite{renyi1959dimension,jalali2014universal}. The mentioned literature also provides practical methods that allow measuring efficiently the compressibility of the process. Finally, Lindenstrauss and Tsukamoto~\cite{lindenstrauss2018rate} have shown how this dimension is related to the metric mean dimension, a notion that emerged in the literature of dynamical systems.

\begin{figure}[t]
     \centering
     \begin{subfigure}[t]{0.48\textwidth}
         \centering
         \includegraphics[width=\textwidth]{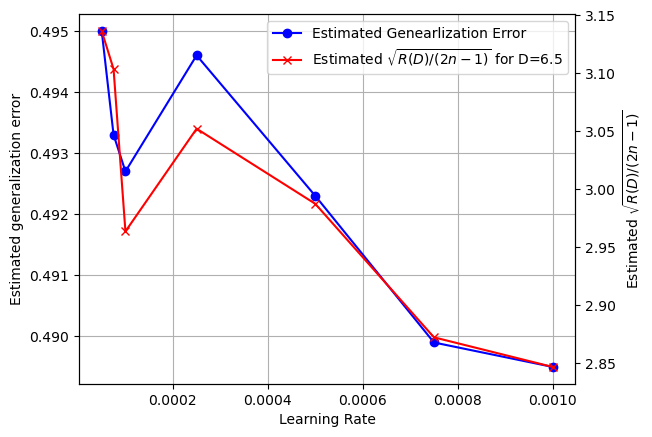}
         \caption{FCN4}
         \label{fig:gen_FCN4}
     \end{subfigure}
          \hfill
          \begin{subfigure}[t]{0.474\textwidth}
         \centering
         \includegraphics[width=\textwidth]{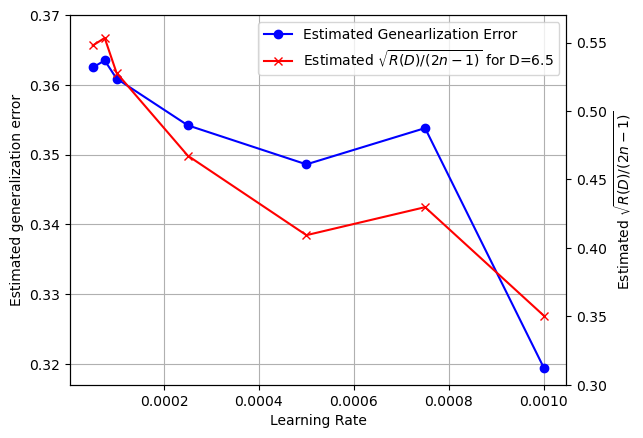}
         \caption{CNN4}
         \label{fig:fen_CNN4}
     \end{subfigure}
        \caption{Estimated generalization error and (an approximation of) the bound of Theorem~\ref{th:RDProcess} computed for FCN4 and CNN4, both trained on the CIFAR10 dataset for various learning rates. The values of the estimated generalization error are plotted relative to the left hand sided Y-axis; and the values of the approximation of the bound of~Theorem~\ref{th:RDProcess} are plotted relative to the right hand sided Y-axis.}
        \label{fig:gen}
\end{figure}

\section{Experiments} \label{sec:experiments}
In this section, we run experiments that validate the result of Theorem~\ref{th:RDProcess}. As also mentioned in~\cite{simsekli2020hausdorff}, the coupling coefficient $M$ is not easy to estimate generally; and, for this reason, similar to in~\cite{simsekli2020hausdorff} hereafter we ignore its contribution to the bound of Theorem~\ref{th:RDProcess}. In our experiments, we follow the following procedure for the classification task with the 0-1 loss function:
\begin{enumerate}
    \item We fix a dataset, a neural network, and the optimization hyperparameters (except the learning rate).
    \item For several learning rates, in the range of $[5e-5,1e-3]$, we train the neural network with the given datasets and normalize and save the optimization trajectories. In doing so, we use the cross-entropy surrogate loss function.
    \item We estimate the generalization errors of the trained networks, each one trained with a different fixed learning rate, using the available ``test set''.
    \item We estimate the rate-distortion values of the optimization trajectories of the trained networks, each one trained with a different fixed learning rate, using ``Neural Estimator of the Rate-Distortion'' (NERD) \cite{Lei22Estimation} and by considering the Root-mean-square error (RMSE) as the distortion function.
\end{enumerate}

\begin{itemize}
\item[] \textbf{Datasets:} We consider image classification using the dataset CIFAR10 \cite{krizhevsky2009learning}. This dataset contains ten classes of color images of size $32 \times 32$. We use $5 \times 10^{4}$ samples for training and  $10^{4}$ samples for the test phase in order to estimate the resulting population risk.
\item[] \textbf{Models:} We consider two models: (i) a fully connected neural network with four hidden layers, which we denote here as FCN4, with a width of $1024$ and total number of parameters of 6,305,802; and (ii)  a concatenation of four convolutional layers and a fully connected network with a hidden layer, which we denote as CNN4, with total number of parameters of 1,065,554. 
\item[] \textbf{Loss function:} We consider the 0-l loss function for the classification task and for the estimation of the generalization error. The neural networks are trained using the cross-entropy surrogate loss function.
\item[] \textbf{Training and hyperparameters:}
Training is performed using SGD with a fixed learning rate ($\eta$) in the range $[5e-5,1e-3]$ and a batch size of $50$. All models are trained until they reach a training accuracy of $100\%$ and a negative training log-likelihood of less than $5e-3$. The models are trained for ten more epochs during which the parameters of the models are first normalized and then stored. We normalize the weights for a fair comparison of the rate-distortion values of distinct models trained with different learning algorithms. These stored models are used as a ``dataset'' in the next step, which measures their lossy compressibility for a given distortion level.
\item[] \textbf{Estimation of rate-distortion values:}
We use the ``Neural Estimator of the Rate-Distortion'' (NERD) \cite{Lei22Estimation} to measure the lossy compressibility of the optimization trajectories. Due to the large size of the trained model, for example for FCN4, the dataset with the dimension 6,305,802, we choose a layer, randomly sample some of the neurons from the input and output of that layer, and choose the corresponding weights between them. Then, we use NERD to measure the compressibility of this smaller-dimension ``dataset''. We repeat this process for each layer, and the reported values are the average of the derived values, scaled to the actual size (dimension) of the model.
\end{itemize}

Additional details on the experiments using the two models can be found in Appendix~\ref{sec:experimentsDetails}. 

Fig~\ref{fig:gen} depicts the evolution of the true (estimated) generalization error for both experiments, as a function of the learning rate. Also shown for comparison, an approximation of the bound of Theorem~\ref{th:RDProcess}. The following observations are in order.
\begin{itemize}
    \item[1.]  The estimate of the bound of Theorem~\ref{th:RDProcess}, which is computed using the rate-distortion estimates of the optimization trajectories, exhibits a similar behavior to that of the true  (estimated) generalization error in terms of the evolution as a function of the learning rate. This means that the degree of compressibility of the optimization trajectory can be used as a good indicator of the generalization capability of the used neural network. 
   \item[2.] Contrasting the values of the (estimated) true generalization of FCN4 against those of CNN4, it is easy to see that the latter generalizes better comparatively. It is insightful to observe that the optimization trajectory of CNN4 is more compressible than that of FCN4. In addition, this holds true for all used learning rates. This observation is consistent with prior art work~\cite{du2018many,li2021why,wang2024theoretical} that reported similar experimental observations on that convolutional neural networks generally generalize better than fully connected ones.
   \item[3.] For both models, it is easy to see that larger learning rates allow smaller generalization errors, an aspect which was also observed, e.g., in~\cite{simsekli2020hausdorff,barsbey2021heavy}. Accordingly, the optimization trajectories are more compressible in that range of learning rates. 
\end{itemize}

\section{Conclusion and future works}
In this work, we have developed a new \emph{unifying variable-size compressibility} framework and have shown how the generalization error of a statistical learning algorithm can be related formally to its degree of `compressibility' as defined in this paper. Furthermore, we have shown how this compressibility approach recovers and possibly subsumes several existing generalization bounds that were previously obtained through seemingly unrelated approaches (and associated proofs), namely rate-distortion-theoretic, information-theoretic, PAC-Bayes, and intrinsic-dimension type bounds. This compressibility framework is further used to establish novel data-dependent upper bounds on the generalization error, such as a new \emph{lossy} PAC-Bayes bound (Proposition~\ref{prop:PAClossy}) and a new intrinsic dimension bound (Theorem~\ref{th:RDProcess}). The latter result formally connects the generalization error of a learning algorithm to the compressibility of its optimization trajectories. This is also validated through some experiments. The result also relates the generalization error to the rate-distortion dimension of a process, the Rényi information dimension of a process, and the metric mean dimension. 

The results of this paper call for several interesting future directions. For example, while classic rate-distortion theory establishes that the communication cost of a given model (i.e., the minimum number of bits that are required to convey it from a given point to another one) essentially depends on the degree of its compressibility, the results of the present paper also relate some form of compressibility (which is variable-size, as we introduce here) to the generalization error of that model. Hence, the framework of variable-size compressibility appears to constitute an interesting common ground for the study of the important question of reconciling (or not ?) the seemingly unrelated requirements of (i) small communication cost and (ii) good generalization capability of statistical learning algorithms. Furthermore, it is noteworthy that the approach of this paper can also be extended in various ways, e.g.,  in terms of the compressibility of the extracted features (also sometimes called latent representations in representation-type learning) or of the output predictions in a manner that is similar to~\cite{sefidgaran2024minimum}. Finally, distributed learning architectures are attracting increasing interest; and our variable-size compressibility is instrumental towards obtaining data-dependent generalization bounds for these settings, as used partially in \cite{sefidgaran2023federated}.

\section*{Acknowledgement}
The authors would like to thank Umut \c{S}im\c{s}ekli and Yijun Wan for some helpful discussions on some aspects of this work.

\bibliographystyle{IEEEtran}
\bibliography{biblio}

\appendices

{\begin{center}	
		\Large{\textbf{Appendices}}		
\end{center}}
The appendices are organized as follows:
\begin{itemize}
\item In Appendix~\ref{sec:limitations}, we show that our rate-distortion theoretic bounds of Theorem~\ref{th:expectation} and Proposition~\ref{prop:PAClossy} do \textit{not} suffer from the limitations of the Xu-Ragisnsky bound as mentioned in~\cite{haghifam2023limitations}.
	\item In Appendix~\ref{sec:HausdorffSGD}, we show how our framework can recover certain previous dimension-based results,
	\item In Appendix~\ref{sec:furtherResults}, a simple corollary of the tail bound is presented.
 \item In Appendix~\ref{sec:experimentsDetails}, we provide further the details about the experiments of Section~\ref{sec:experiments}.
	\item Appendix~\ref{sec:proofs} contains the proofs.
\end{itemize}


\section{\texorpdfstring{$\mathcal{O}(1/n)$}{O(1/n)} Generalization Bounds for the  Setup of~\texorpdfstring{\cite{haghifam2023limitations}}{[59]}}  \label{sec:limitations}

Despite the popularity of PAC-Bayes and information-theoretic generalization bounds, their relevance and utility in practice are sometimes questioned. The recent works~\cite{haghifam2023limitations} and~\cite{livni2023information} argue that such bounds are not expressive enough generally and fail to give non-vacuous bounds in some cases. In what follows we show that the results of rate-distortion theoretic bounds of Theorem~\ref{th:expectation} and Proposition~\ref{prop:PAClossy} escape those limitations. Specifically, for the setup considered in~\cite{haghifam2023limitations} we obtain in-expectation and PAC-Bayes generalization bounds for the considered that are not only non-vacuous but decay with $n$ as $\mathcal{O}(1/n)$.

We start by recalling the counter-example setup used in ~\cite{haghifam2023limitations} to infer that information-theoretic type bounds such as \cite{russozou16,xu2017information} do fall short to explain generalization and expressiveness. The authors of~\cite{haghifam2023limitations} also provide a similar counter-example and arguments for conditional mutual information (CMI)-type bounds~\cite{steinke2020reasoning}. However, for reasons of brevity and compatibility with our framework, we do not discuss them hereafter.

\subsection{Counter-example of~\texorpdfstring{\cite{haghifam2023limitations}}{[59]}} \label{sec:limitations_setup}
Consider the following stochastic convex optimization (SCO) problem optimized using Gradient Descent (GD) as stated in~\cite{haghifam2023limitations}.

\textbf{Constants:} Fix an $n\in \mathbb{N}$. Let $T_n = 2n^2$, $\eta_n = \frac{1}{n\sqrt{5n}}$, $d_n = \frac{3}{4}T_n 2^n$, and $\lambda_n =\frac{1}{n\sqrt{d}}$. In the rest of the section, we drop the dependencies (subscripts) on $n$ for better readability.

\textbf{Data distribution and training set.} Let $\mathcal{Z} =\{0,1\}^d$. Let $\vc{Z}=(Z_1,\ldots,Z_d)\in \mathcal{Z}$ be a random vector whose elements are i.i.d. and distributed according to  the Bernoulli$(1/2)$ distribution, \ie $\vc{Z}\sim \text{Bern}(1/2)^{\otimes d} \eqqcolon \mu$. Denote the elements of the training dataset $S$ of size $n$ by 
\begin{align*}
    S =\{\vc{Z}_1,\vc{Z}_2,\ldots,\vc{Z}_n\}\sim \mu^{\otimes n}, \quad \quad \vc{Z}_i=(Z_{i,1},Z_{i,2},\ldots,Z_{i,d}) \in \mathcal{Z}, i\in[n].
\end{align*}

\textbf{Hypothesis space and loss function.} The hypothesis space $\mathcal{W}$ is the ball of radius 1 in $\mathbb{R}^d$ and the considered loss function  $\ell\colon \mathcal{Z} \times \mathcal{W} \to \mathbb{R}^+$ is defined as
\begin{align}
    \ell(\vc{z},\vc{w}) \coloneqq \sum_{j\in[d]} z_j w_j^2 + \lambda \langle \vc{w} , \vc{z} \rangle + \max \left\{ \max_{j\in[d]} \left\{w_j\right\},0\right\},
    \label{choice-loss-function-counter-example}
\end{align}
where $\vc{z}=(z_1,z_2,\ldots,z_d)\in \mathcal{Z}$ and $\vc{w}=(w_1,w_2,\ldots,w_d)\in \mathcal{W}$.

\textbf{Learning algorithm.} The considered learning algorithm $\mathcal{A}\colon \mathcal{Z}^n \to \mathcal{W}$ is the output of the $T$ rounds of iterations of the Gradient Decent (GD) algorithm, initialized by $\vc{W}_0 = \vc{0} \in \mathbb{R}^d$. More precisely, for $t \in [T]$, let
\begin{align*}
    \vc{W}^{\{t+1\}} = \Pi_{\mathcal{W}} \left(\vc{W}^{\{t\}}-\eta  \nabla \hat{\mathcal{L}}(S,\vc{W}^{\{t\}})\right), 
\end{align*}
where $\hat{\mathcal{L}}(S,\vc{W}^{\{t\}})=\frac{1}{n}\sum_{i\in[n]} \ell(\vc{Z}_i,\vc{W}^{\{t\}})$ and $ \Pi_{\mathcal{W}}\colon \mathbb{R}^d \to \mathcal{W}$ denotes the Euclidean projection operator, \ie
\begin{align*}
    \Pi_{\mathcal{W}}(\vc{x}) \coloneqq \argmin_{\vc{W}\in \mathcal{W}} \|\vc{W}-\vc{x}\|_2 = \frac{\vc{x}}{\max(\|\vc{x}\|_2,1)}.
\end{align*} 
With the above notations, $\mathcal{A}(S)=\vc{W}^{\{T\}}$.

\subsection{\texorpdfstring{$\mathcal{O}(1/n)$}{O(1/n)} Generalization Gap Bounds}

In~\cite{haghifam2023limitations}, the authors show that the Xu-Raginsky information-theoretic upper bound on the generalization error of~\cite{xu2017information} is loose and falls short to characterize the behavior of the generalization error for the aforementioned counter-example. This is obtained by showing that: (i) for every $\sigma > 0$ it holds that~\cite[Theorem 4]{haghifam2023limitations} 
\begin{align}
   \sqrt{\frac{2 I(S;\hat{\vc{W}})}{n}}+ \mathbb{E}_S\left[\left|\mathcal{L}\left(\hat{\vc{W}}\right)-\mathcal{L}\left(\vc{W}^{\{T\}}\right)\right|\right]+\mathbb{E}_S\left[\left|\hat{\mathcal{L}}\left(S,\hat{\vc{W}}\right)-\mathcal{S}\left(S,\vc{W}^{\{T\}}\right)\right|\right] = \Omega(1)
\end{align}
where $\hat{\vc{W}} = \Pi_{\mathcal{W}} \left( \vc{W}^{\{T\}} + \xi \right)$, and $\xi \sim \mathcal{N}\left(0,\sigma^2 \mathrm{I}_d\right)$; and (ii) that the true in-expectation generalization error satisfies
\begin{align}
    \mathbb{E}_{S}\left[\gen(S,\vc{W}^{\{T\}})\right] = \mathcal{O}\left(1/\sqrt{n}\right).
    \label{bound-generalization-error-counter-example}
\end{align}
 The authors also observe that considering a perturbation of the bound of~\cite{xu2017information} with an i.i.d. Gaussian noise $\xi$ does not resolve the issue. Furthermore, in~\cite[Theorem~9]{haghifam2023limitations} they establish that when applied to the same counter-example the PAC-Bayes bounds of~\cite{mcallester1998some,mcallester1999pac} suffer from similar limitations.
 
In what follows, we show that our rate-distortion theoretic results of Theorem~\ref{th:expectation} and Proposition~\ref{prop:PAClossy} lead to in-expectation and PAC-Bayes generalization bounds for the counter-example that are not only non-vacuous (by opposition to that of Xu-Raginsky~\cite{xu2017information}) but are tighter than the right-hand side of~\eqref{bound-generalization-error-counter-example}. Specifically, application of our Theorem~\ref{th:expectation} yields that
\begin{align}
    \mathbb{E}_S\left[\gen\left(S,\vc{W}^{\{T\}}\right)\right] = \mathcal{O}\left(1/n\right);
     \label{in-expectation-bound-generalization-error-counter-example}
\end{align}
and application of our Proposition~\ref{prop:PAClossy} yields that for any fixed $\delta \geq 0$ (that does not grow with $n$), with probability at least $1-\delta$ over the choice of $S$, we have
    \begin{align}
    \gen\left(S,\vc{W}^{\{T\}}\right) = \mathcal{O}\left(1/n\right).
    \label{tail-bound-generalization-error-counter-example}
\end{align}
The proofs of~\eqref{in-expectation-bound-generalization-error-counter-example} and~\eqref{tail-bound-generalization-error-counter-example} can be found in Appendix~\ref{pr:example}.


\section{Relation to dimension-based bounds} \label{sec:HausdorffSGD}
In this section, we show how using Theorem~\ref{th:generalTail} we can recover certain dimension-based bounds, and in particular, \cite[Theorem~2]{simsekli2020hausdorff}. We state the assumptions used in \cite[Theorem~2]{simsekli2020hausdorff} in their simplest form. We refer the readers to \cite{simsekli2020hausdorff} for further precision.  

{\c S}im{\c s}ekli et al.~\cite{simsekli2020hausdorff} considered the continuous-time model of the SGD. Suppose that $U\in \mathcal{U}$ is the stochasticity of the algorithm, which is independent of $S$. Denote the deterministic output of the algorithm at time $t$, when $S=s$ and $U=u$, as $W_{s|u}^{t}\coloneqq \mathcal{A}(s,t|u) \in \mathbb{R}^d$. Denote $W_{s}^{t}\coloneqq W_{s|U}^{t}$, as a random variable representing the model at time $t$ for a given $s$, that depends on $U$. For any $u$ and $s$, denote $\mathcal{W}_{s|u}\coloneqq\{ w \in \mathcal{W}\colon \exists  t \in \T, w=\mathcal{A}(s,t|u)\}$ as the support set of the trajectory of the hypotheses for a particular dataset $s$ and the stochasticity $u$. To recall, we assume the interval as $\T\coloneqq [0,1] \subseteq \mathbb{R}$ and denote the set of hypotheses along the trajectories by $W^{\T}$. Denote $\mathcal{W}_s\coloneqq\mathcal{W}_{s|U}$, as a random variable representing the random trajectory for a given $s$. To state their result, consider the collection of closed balls of radius $\epsilon$, with the centers on the fixed grid
\begin{align*}
	\mathcal{N}_{\epsilon}\coloneqq \left\{\left(\frac{(2j_1+1)\epsilon}{2\sqrt{d}},\cdots,\frac{(2j_d+1)\epsilon}{2\sqrt{d}}\right)\colon j_i \in \mathbb{Z},i=1,\ldots,d\right\}.
\end{align*}
For each $u$ and $s$, define the set  $\mathcal{N}_{\epsilon}(s|u)\coloneqq \{x\in \mathcal{N}_{\epsilon}\colon B_d(x,\epsilon)\bigcap \mathcal{W}_{s|u}\neq \varnothing \}$,  where $B_d(x,\epsilon)$ denotes the closed ball centered around $x \in \mathbb{R}^d$ with radius $\epsilon$. The following assumption is mainly used in their works.
\begin{assumption}[{\cite[Definition~5]{simsekli2020hausdorff}}] \label{a:M}
	Fix some $u \in \mathcal{U}$. Let $\mathcal{Z}^{\infty}\coloneqq(\mathcal{Z}\times \cdots \times \mathcal{Z})$ denote the countable product endowed with the product topology and let $\mathfrak{B}$ be the Borel $\sigma$-algebra generated by $\mathcal{Z}^{\infty}$.  Let $\mathfrak{F},\mathfrak{G}$ be the $\sigma$-subalgebras of $\mathfrak{B}$ generated by the collection of random variables given by $\{\mathcal{\hat{L}}(s,w)\colon w \in \mathcal{W}, n\geq 1\}$ and $\left\{\bm{1}\{w \in \mathcal{N}_{\epsilon}(s|u)\}\colon \delta\in \mathbb{Q}_{>0},w \in \mathcal{N}_{\epsilon}, n\geq 1\right\}$. 
	There exists a constant $M\geq 1$ such that for any $F\in \mathfrak{F}$ and $G \in \mathfrak{G}$, we have $\mathbb{P}(F \bigcap G) \leq M\mathbb{P}(F) \mathbb{P}(G)$.
\end{assumption}
The following result makes a connection between the generalization error and the \emph{Hausdorff} dimensions ($\dim_H$) of the optimization trajectories. The result is established when this dimension coincides with the \emph{Minkowski} dimension ($\dim_M$). For brevity, we refer the readers to \cite[Definitions~2 \& S3]{simsekli2020hausdorff} for the definition of the Hausdorff and Minkowski dimensions.\footnote{Here, we state the result for 0-1 loss function, that can be trivially extended for any $B$ bounded loss.}

\begin{theorem}[{\cite[Theorem~2]{simsekli2020hausdorff}}] Suppose that the loss $\ell\colon \mathcal{Z}\times \mathcal{W}\to [0,1]$ is $\mathfrak{L}$-Lipschitz. Suppose that almost surely $\dim_H(W_{S|U})=\dim_M(W_{S|U})<\infty$. Then, under Assumption~\ref{a:M}, with probability at least $1-\delta$,
	\begin{align*}
		\sup_{t\in[0,1]} |\gen(S,W^t_{S})|{=}\sup_{W \in \mathcal{W}_{S}} |\gen(S,W)|\leq 2 \sqrt{\frac{(\dim_H(\mathcal{W}_{S|U})+1)\log^2(n\mathfrak{L}^2)}{n}+\frac{\log(7M/\delta)}{n}}.
	\end{align*}
\end{theorem}

Now, we prove this result, using Theorem~\ref{th:generalTail}.i and more precisely using Corollary~\ref{cor:generalTailU}, that takes into account the stochasticity $U$.
\begin{IEEEproof} For each $s$ and $u$, denote the size of the set $\mathcal{N}_{\epsilon}(s|u)$ as $N_{\epsilon}(s|u) \coloneqq |\mathcal{N}_{\epsilon}(s|u)|$, which is bounded due to Assumption H2 of \cite{simsekli2020hausdorff}. As it is shown in \cite{simsekli2020hausdorff}, for any $\delta'>0$, there exists a set $\tilde{\mathcal{S}} \subseteq \mathcal{S}$ with probability at least $1-\delta'$ such that for this set, $\epsilon\coloneqq 1/(n\mathfrak{L}^2)$, and sufficiently large $n$, 
	\begin{align}
		\log N_{\epsilon}(S|U) \leq (\dim_H(\mathcal{W}_{S|U})+1) \log^2(n\mathfrak{L}^2). \label{eq:HausdorffNumber}
	\end{align}
	Note that the results used in \cite{simsekli2020hausdorff} are for $\epsilon\coloneqq 1/\sqrt{n\mathfrak{L}^2}$. However, their general expression for bounding $N_{\epsilon}(S|U) $ is not limited to that choice. Hence,
	\begin{align}
		\mathbb{P}&\left(\sup_{W \in \mathcal{W}_{S}} |\gen(S,W)|>2\sqrt{\frac{(\dim_H(\mathcal{W}_{S|U})+1)\log^2(n\mathfrak{L}^2)}{n}+\frac{\log(7M/\delta)}{n}}\right) \nonumber\\
		&\stackrel{(a)}{\leq} \mathbb{P}\left(\sup_{W \in \mathcal{W}_{S}} |\gen(S,W)|>\sqrt{\frac{(\dim_H(\mathcal{W}_{S|U})+1)\log^2(n\mathfrak{L}^2)+\log(4Mn/\delta)}{2n-1}+4\mathfrak{L}\epsilon}\right)\nonumber\\
		&\leq\delta'+\mathbb{P}\left(\sup_{W \in \mathcal{W}_{S}} |\gen(S,W)|>\sqrt{\frac{\log N_{\epsilon}(S|U)+\log(4Mn/\delta)}{2n-1}+4\mathfrak{L}\epsilon}\right), \label{eq:HausdorffNumber2}
	\end{align}
	where $(a)$ holds for $n$ sufficiently large and for $\epsilon\coloneqq 1/(n\mathfrak{L}^2)$.
	
	Let $\delta'=\delta/2$. We show that the last term of \eqref{eq:HausdorffNumber2} is bounded by $\delta/2$, which completes the proof. To do so, we use Corollary~\ref{cor:generalTailU}.i. and let
	$w \mapsto w^{\T}$ and $\delta \mapsto \delta/2$, $\epsilon \mapsto 4\mathfrak{L}\epsilon$,
	\begin{align*}
		f(s,w^{\T})\coloneqq \left(\sup_{t\in [0,1]} \gen(s,w^t_s)\right)^2,\quad \Delta(u,s)\coloneqq \frac{\log N_{\epsilon}(s|u)+\log(4Mn/\delta)}{2n-1}+4\mathfrak{L}\epsilon. 
	\end{align*}
	Let $\hat{\mathcal{W}} \coloneqq \mathcal{W}$ and $g(s,\hat{w})\coloneqq \gen(s,\hat{w})^2$.  	Note that $\forall w,\hat{w} \in \mathcal{W}$, we have $|\gen(s,w)-\gen(s,\hat{w})|\leq 4\mathfrak{L}\|w-\hat{w}\|$. Thus, for any given $s$ and $u$, and any $\nu_{u,s,w}\in \mathcal{F}_{U,S,W^{\T}}^{\delta}$, there exists at least one $\hat{w} \in \mathcal{N}_{\epsilon}(s|u)$, such that 
	\begin{align*}
		g(s,\hat{w})+4\mathfrak{L}\epsilon \geq f(s,w_{[0:1]}) \geq \Delta(u,s),
	\end{align*}
	where the last inequality is deduced since $\nu_{u,s,w^t} \in \mathcal{F}^{\delta}_{U,S,W^{\T}}$.  For any $s$ and $u$, denote such a set of $\hat{w}$ such that $g(s,\hat{w})+4\mathfrak{L}\epsilon \geq \Delta(u,s)$ as $\mathcal{N}^*_{\epsilon}(s|u)$. 
	Then, let $p_{\hat{W}|U,S}$ be a deterministic (delta Dirac) distribution, picking one of the  $\hat{w}\in \mathcal{N}^*_{\epsilon}(s|u)$. This choice of distribution satisfies the distortion criterion \eqref{eq:tailDistortionU} with $\epsilon \mapsto 4\mathfrak{L}\epsilon$. Note that $\hat{W}$ and $W^{\T}$ are not of the same dimension. Moreover, $\hat{W}$ is a discrete random variable and deterministic given an $S=s$ and $U=u$. For any $s$ and $u$, let $q_{\hat{W}|U,S}$ be a uniform distribution over $\mathcal{N}^*_{\epsilon}(s|u)$. Now, for any $\nu_{u,s,w^t} \in \mathcal{F}^{\delta}_{U,S,W^{\T}}$,
	\begin{align}
		D_{KL}\left(p_{\hat{W}|U,S} \nu_{U,S}\|q_{\hat{W}|U,S} \nu_{U,S}\right) =&\mathbb{E}_{\nu_{U,S}}[\log N_{\epsilon}(S|U)]. \label{eq:HausdorffNumber3}
	\end{align}
	Furthermore, with $\lambda \coloneqq (2n-1)$,
	\begin{align}
		\log\mathbb{E}_{P_UP_S q_{\hat{W}|S}}\left[e^{\lambda g(S,\hat{W})}\right]\stackrel{(a)}{\leq}& \log \left(M\mathbb{E}_{P_U P_S q_{\hat{W}}}\left[e^{\lambda \gen(S,\hat{W})^2}\right]\right) \nonumber \\ 	\leq& \log(M)+\log \left(2n\right). \label{eq:HausdorffNumber4}
	\end{align}
	Relations \eqref{eq:HausdorffNumber3} and \eqref{eq:HausdorffNumber} conclude that Condition~\eqref{eq:ConditionRenyiU} is satisfied for any $\nu_{u,s,w}$. This completes the proof.
\end{IEEEproof}

Another notable work showing a connection between generalization error and the intrinsic dimensions is by Camuto et al.~\cite{camuto2021fractal}. They have studied the generalization error of the stochastic learning algorithms that can be expressed as \emph{random iterated function systems} (IFS). In this case, and under mild conditions, they made a connection between the generalization error and the Hausdorff dimension of the induced invariant measure given $S$. Similar to the approach above, one can recover \cite[Theorem~1]{camuto2021fractal}. Here, we avoid showing the steps, as heavy notations and definitions are needed to introduce the setup of that work.

\section{A corollary of the tail bound} \label{sec:furtherResults}
As mentioned, if any random variable $U$ representing the (partial) stochasticity of the algorithm is known, the bound may be improved by making the sufficient conditions conditioned on $U$, in a similar way as in \cite[Section D.2, Theorem~10]{Sefidgaran2022}. Here, we state the result, whose proof follows from Theorem~\ref{th:generalTail}, by letting $S \mapsto U$ and $P_S \mapsto P_S P_U$. 
\begin{corollary} \label{cor:generalTailU} Suppose that $U\sim P_U$ represents the (partial) stochasticity of the algorithm, that is independent of $S$, \ie the joint distribution of $(U,S,W)$ can be written as $P_{U,S,W}=P_U P_S P_{W|S,U}$. Let $f(S,W)\colon \mathcal{U} \times \mathcal{S} \times \mathcal{W} \to \mathbb{R}$ and $\Delta(U,S,W)\colon \mathcal{U}\times \mathcal{S} \times \mathcal{W} \to \mathbb{R}^+$.  Fix arbitrarily the set $\hat{\mathcal{W}}$ and define arbitrarily $g(S,\hat{W})\colon \mathcal{U} \times \mathcal{S} \times \mathcal{\hat{W}} \to \mathbb{R}$. Then, for any $\delta \in \mathbb{R}^+$, with probability at least $1-\delta$,  
	\begin{align}
		f(S,W)\leq \Delta(U,S,W), \label{eq:tailU}
	\end{align}
	if any of these conditions hold:
	\begin{itemize}[leftmargin=*]
		\item[i.] For some  $\epsilon \in \mathbb{R}$ and any $\nu_{U,S,W} \in \mathcal{F}_{U,S,W}^{\delta}\coloneqq \mathcal{G}_{U,S,W}^{\delta}\bigcap\mathcal{S}_{U,S,W}\left(f(s,w)- \Delta(u,s,w)\right)$,\footnote{$\mathcal{G}_{U,S,W}^{\delta}$, $\mathcal{S}_{U,S,W}(\cdot)$, and $\mathfrak{T}_{\alpha,P_UP_S}(\nu_{U,S},p_{\hat{W}|S},q_{\hat{W}|S},\lambda f)$ are defined in \eqref{def:Gdelta}, \eqref{def:fSupport}, and \eqref{def:T},  respectively.}
		\begin{align}
			\inf_{p_{\hat{W}|U,S}\in \mathcal{Q}(\nu_{U,S,W})}\inf_{\lambda > 0,q_{\hat{W}|U,S}}  \bigg\{&\mathfrak{T}_{1,P_U P_S}(\nu_{U,S},p_{\hat{W}|U,S},q_{\hat{W}|U,S},\lambda g)\nonumber \\
			&-D_{KL}(\nu_{U,S,W}\|P_{W|U,S}\nu_{U,S}) \nonumber \\
			&{-}\lambda \left( \mathbb{E}_{\nu_{U,S,W}}\left[\Delta(U,S,W)\right]-\epsilon\right)\bigg\} \leq \log(\delta),\label{eq:ConditionRenyiU}
		\end{align}
		where $\nu_{U,S}$ and $\nu_{S,W}$ are the marginal distributions of $(U,S)$ and $(S,W)$, respectively, under $\nu_{U,S,W}$ and $\mathcal{Q}(\nu_{U,S,W})$ is the set of all distributions $p_{\hat{W}|U,S}$ such that 
		\begin{align}
			\mathbb{E}_{\nu_{U,S,W} p_{\hat{W}|U,S}}\left[\Delta(U,S,W)-g(S,\hat{W})\right] \leq \epsilon. \label{eq:tailDistortionU}
		\end{align}
		\item[ii.] For some $\alpha>1$ and any $\nu_{U,S,W} \in \mathcal{F}_{U,S,W}^{\delta}$,
		\begin{align}
			\inf_{q_{W|U,S},\lambda \geq \alpha/(\alpha-1)}  \bigg\{&\mathfrak{T}_{\alpha,P_UP_S}(\nu_{U,S},\nu_{W|U,S},q_{W|U,S},\lambda \log(f)) -\lambda \left( \mathbb{E}_{\nu_{U,S}}\log \mathbb{E}_{\nu_{W|U,S}}\left[\Delta(U,S,W)\right]\right)\bigg\} \leq \log(\delta).\label{eq:ConditionRenyiImprovedU}
		\end{align}
	\end{itemize}
\end{corollary}

\section{Additional Details of the Experiments of Section~\ref{sec:experiments}} \label{sec:experimentsDetails}

We performed our experiments on a server that is equipped with 56 CPUs Intel Xeon E5-2690v4 2.60GHz and 4 GPUs Nvidia Tesla P-100 PCIe 16GB. Our implementation uses the Python language and the deep learning framework PyTorch \cite{paszke2019pytorch}.

 As for the models, using the torch.nn submodule of PyTorch the architecture of FCN4 can be described as 
\begin{align*}
    \text{nn.Sequential}\Big(&  \text{nn.Linear}(3072, 1024),              \text{nn.LeakyReLU}(),          \text{nn.Linear}(1024, 1024),          \text{nn.LeakyReLU}(),\\
       &         \text{nn.Linear}(1024, 1024),        \text{nn.LeakyReLU}(),      \text{nn.Linear}(1024, 1024),     \text{nn.LeakyReLU}(),   \text{nn.Linear}(1024, 10)\Big)
\end{align*}
Similarly, the architecture of CNN4 can be described as
\begin{align*}
    \text{nn.Sequential}\Big( &  
     \text{nn.Conv2d}(3, 8, 5, 1, 2),\text{nn.Conv2d}(8, 8, 5, 1, 2),\text{nn.LeakyReLU}(),\text{nn.MaxPool2d}(2, 2),\\
    &\text{nn.Conv2d}(8, 16, 3, 1, 1),
                \text{nn.Conv2d}(16, 16, 3, 1, 1),
                \text{nn.LeakyReLU}(),
                \text{nn.MaxPool2d}(2, 2),
               \\
            &  \text{nn.Flatten}(),\text{nn.Linear}(1024, 1024),     \text{nn.LeakyReLU}(),    \text{nn.Linear}(1024, 10) \Big).
\end{align*}


\section{Proofs} \label{sec:proofs}
In this section, we provide the proofs of the results of the paper.


 \subsection{Proof of Theorem~\ref{th:compressibility}} \label{pr:compressibility}
For ease of notations, let $\mathcal{E}$ be the event that $\min_{j \leq e^{\sum_{i\in [m]}R_{S_i,W_i}}} d_m(W^m,\hat{\vc{w}}[j];S^m) >\epsilon$ and $\mathcal{E}^C$ be its complement. For any $\nu \in (0,\log(1/\delta))$, let $m_0$ be sufficiently large such that for any $m\geq m_0$, $\mathbb{P}_{(S,W)^{\otimes m}}\left(\mathcal{E}\right)\leq e^{-m(\log(1/\delta)-\nu)}$. Let $\Delta(S,W)\coloneqq \frac{4\sigma^2(R_{S,W}+\log(\sqrt{2n}/\delta))}{2n-1}+\epsilon$. 
	Then,
	\begin{align}
		\mathbb{P}_{(S,W)}&\Big(\gen(S,W)> \sqrt{\Delta(S,W)}\Big)^m\nonumber  \\ &=\mathbb{P}_{(S,W)^{\otimes m}}\Big(\forall i \in [m]\colon \gen(S_i,W_i)^2
		> \Delta(S_i,W_i) \Big)\nonumber \\ 
		&\leq \mathbb{P}_{(S,W)^{\otimes m}}\bigg(\frac{1}{m}\sum_{i\in [m]}\gen(S_i,W_i)^2
		> \frac{1}{m}\sum_{i\in [m]}\Delta(S_i,W_i) \bigg)\nonumber \\
		&\leq \mathbb{P}_{(S,W)^{\otimes m}}\bigg(\frac{1}{m}\sum_{i\in [m]}\gen(S_i,W_i)^2
		> \frac{1}{m}\sum_{i\in [m]}\Delta(S_i,W_i) ,\mathcal{E}^C\bigg)+\mathbb{P}_{(S,W)^{\otimes m}}(\mathcal{E})\nonumber \\
		&\leq \mathbb{P}_{(S,W)^{\otimes m}}\bigg(\frac{1}{m}\sum_{i\in [m]}\gen(S_i,W_i)^2
		> \frac{1}{m}\sum_{i\in [m]}\Delta(S_i,W_i) ,\mathcal{E}^C\bigg)+e^{-m(\log(1/\delta)-\nu)}\nonumber \\
		&\leq \mathbb{P}_{S^{\otimes m}}\bigg(\exists j\leq e^{\sum_{i\in [m]}R_{S_i,W_i}}\colon \frac{1}{m}\sum_{i\in [m]}\gen(S_i,\hat{w}_i[j])^2
		> \frac{1}{m}\sum_{i\in [m]}\Delta(S_i,W_i) -\epsilon \bigg)\nonumber\\
		&\hspace{0.5 cm}+e^{-m(\log(1/\delta)-\nu)} \nonumber\\
		&\stackrel{(a)}{\leq} 	\delta^m \log(e^{m R_{\max}+m\,\eta_m})+e^{-m(\log(1/\delta)-\nu)}\nonumber \\
		&= \delta^m (m R_{\max}+m\,\eta_m)+e^{-m(\log(1/\delta)-\nu)}, \label{eq:prComp1}
	\end{align} 
	where $(a)$ is derived in the following for some sequence of $\eta_m$ such that $\lim_{m\ra\infty} \eta_m =0$. The proof completes by taking the $m$'th root, and since $\nu$ can be chosen arbitrarily small.
	
	Now, we show the step $(a)$:
	\begin{align}
		\mathbb{P}_{S^{\otimes m}}&\left(\exists j  \leq e^{\Sigma_i R_{S_i,W_i}}\colon \sum_i \gen(S_i,\hat{w}_{i}[j])^2> 4 \sigma^2\sum_i\frac{R_{S_i,W_i}+\log(\sqrt{2n}/\delta)}{2n-1}\right)\nonumber\\
		&= \mathbb{P}_{S^{\otimes m}}\left(\exists j  \leq e^{\Sigma_i R_{S_i,W_i}}\colon \sum_i \gen(S_i,\hat{w}_{i}[j])^2> 4 \sigma^2\frac{\sum_i R_{S_i,W_i}+m\log(\sqrt{2n}/\delta)}{2n-1}\right)\nonumber\\
		& \leq \mathbb{P}_{S^{\otimes m}}\left(\exists R \leq m R_{\max} \colon \exists j \leq e^{R}\colon \sum_i \gen(S_i,\hat{w}_{i}[j])^2> 4 \sigma^2\frac{R+m\log(\sqrt{2n}/\delta)}{2n-1}\right)\nonumber\\
		& \leq \mathbb{P}_{S^{\otimes m}}\left(\exists R \leq m R_{\max} \colon \exists j \leq e^{R} \colon \sum_i \gen(S_i,\hat{w}_{i}[j])^2> 4 \sigma^2\frac{\log(j)+m\log(\sqrt{2n}/\delta)}{2n-1}\right)\nonumber\\
		& =\mathbb{P}_{S^{\otimes m}}\left( \exists j \leq e^{m R_{\max}}\colon \sum_i \gen(S_i,\hat{w}_{i}[j])^2> 4 \sigma^2\frac{\log(j)+m\log(\sqrt{2n}/\delta)}{2n-1}\right)\nonumber\\
		& \leq \sum_{j \leq e^{m R_{\max}}} \mathbb{P}_{S^{\otimes m}}\left( \sum_i \gen(S_i,\hat{w}_{i}[j])^2> 4 \sigma^2\frac{\log(j)+m\log(\sqrt{2n}/\delta)}{2n-1}\right)\nonumber\\
		&\stackrel{(a)}{\leq} \sum_{j \leq e^{m R_{\max}}} e^{-\left(\log(j)+m\log(\sqrt{2n}/\delta)\right)} \mathbb{E}\left[e^{\frac{\sum_i \gen(S_i,\hat{w}_{i}[j])^2}{4 \sigma^2/(2n-1)}}\right]\nonumber\\
		&\stackrel{(b)}{\leq} \delta^m \sum_{j \leq e^{m R_{\max}}} \frac{1}{j}\nonumber\\
		&\stackrel{(c)}{\leq} \delta^m \log(e^{m R_{\max}+m\,\eta_m}),
	\end{align}
 	where: $(a)$ follows using the Chernoff bound; $(b)$ follows by using \cite{wainwright_2019} and the fact that for each $i\in [m]$, $\gen(S_i,\hat{w}_{i}[j])$ is $\sigma/\sqrt{n}$ subgaussian, which by using \cite[Theorem~2.6.IV.]{wainwright_2019} yields that for any $i\in [m]$ and $\lambda \in [0,1)$, $$\mathbb{E}\left[e^{\frac{\lambda \gen\left(S_i,\hat{w}_{i}[j]\right)^2}{(2\sigma^2/n)}}\right]\leq \frac{1}{\sqrt{1-\lambda}}$$ 
and, so, by letting $\lambda =\frac{2n-1}{2n}$, we get
$$\mathbb{E}\left[e^{\frac{\gen\left(S_i,\hat{w}_{i}[j]\right)^2}{(4\sigma^2/(2n-1))}}\right]=\mathbb{E}\left[e^{\frac{\lambda \gen\left(S_i,\hat{w}_{i}[j]\right)^2}{(2\sigma^2/n)}}\right] \leq \frac{1}{\sqrt{1-\lambda}}=\sqrt{2n};$$ 
and $(c)$ holds for some  sequence of $\eta_m$ such that $\lim_{m\ra\infty} \eta_m =0$ since the \emph{Harmonic series $H_r$} can upper bounded by $\log(r)+0.6+1/2r$.


 \subsection{Proof of Theorem~\ref{th:covering}} \label{pr:covering}
First, we give a proof of the result for the following condition which is more stringent than~\eqref{eq:ConditionRenyiSimp}:
\begin{align}
	\inf_{p_{\hat{W}|S}\in \mathcal{Q}(\nu_{S,W})}\inf_{q_{\hat{W}}}  D_{KL}\left(p_{\hat{W}|S} \nu_{S}\|q_{\hat{W}} \nu_{S}\right)  \leq \mathbb{E}_{\nu_{S,W}}[R_{S,W}].\label{eq:ConditionStrong}
\end{align}
Then we show that Condition~\ref{eq:ConditionRenyiSimp} is sufficient.

Let $Q_m$ be the empirical distribution of a sequence $(s^m,w^m)\in \mathcal{S}^m \times \mathcal{W}^m$; denoted as $\mathcal{T}_m(s^m,w^m)=Q_m$. This is known in information theory as the ``type'' of the sequence. Let 
	\begin{align*}
		\mathcal{Q}_m \coloneqq \left\{Q_m\colon D_{KL}(Q_m \|P_{S,W})\leq \log(1/\delta)\right\}.
	\end{align*}	
	For any $m$, we first start by constructing proper hypothesis books $\mathcal{H}_{m,Q_m}$ for $Q_m \in \mathcal{Q}_m$, with elements $\mathcal{H}_{m,Q_m}=\{\hat{\vc{w}}_{Q_m}[j], j\in [l_{Q_m}]\}$ and the following property for $m$ sufficiently large: for every sequence $(s^m,w^m)$ such that $\mathcal{T}_m(s^m,w^m)=Q_m$, there exists at least one index $j\in [l_{Q_m}]$, such that $d_m(w^m,\hat{\vc{w}}_{Q_m}[j];s^m) \leq \epsilon+\nu_2$. Due to \cite[Section 6.1.2, Lemma 1]{Berger1975} (adapted for leaning algorithm setup in \cite[Lemma~28]{Sefidgaran2022}), for any $p_{\hat{W}|S}$ and $q_{\hat{W}}$ and $m$ sufficiently large, such hypothesis books exist with the size $l_{Q_m}=e^{mD_{KL}(p_{\hat{W}|S} Q_{m,S}\|q_{\hat{W}} Q_{m,S})}$, where $Q_{m,S}$ here denotes the marginal distribution of $S$ under $Q_{m,S}$, \ie the empirical distribution of $s^m$.
	
	We construct the hypothesis book $\mathcal{H}_m$ as $\mathcal{H}_m  \coloneqq \bigcup_{Q_m \in \mathcal{Q}_m} \mathcal{H}_{m,Q_m} \bigcup \{\hat{\vc{w}}_0\}$, where $\hat{\vc{w}}_0\in \mathcal{\hat{W}}^m$ is an arbitrary (dummy) vector. We now \emph{index} elements of $\mathcal{H}_m$ in a proper manner to satisfy the compressibility condition \eqref{eq:condComp}. Let $r\coloneqq |\mathcal{Q}_m|$, which is upper bounded by $m^{|\mathcal{S}|\times |\mathcal{W}|}$, and consider an arbitrary ordering $\pi(Q_m)\colon \mathcal{Q}_m \to [r]$ (with its inverse denoted as $\pi^{-1}(i)$, $i\in [r]$) for elements of $Q_m \in \mathcal{Q}_m$. Then, we let $\mathcal{H}_{m}=\{\hat{\vc{w}}[r j_1+j_2],j_2\in [r-1],j_1\leq e^{m R_{\max}}/r-1 \}$, where $\hat{\vc{w}}[rj_1+j_2]\coloneqq \hat{\vc{w}}_{\pi^{-1}(j_2)}[j_1]$, if $|\pi^{-1}(j_2)|\leq j_1$, and otherwise $\hat{\vc{w}}[rj_1+j_2]\coloneqq \hat{\vc{w}}_{0}$. Hence, for any distribution $Q_m$, it is sufficient to consider only the first $r\times |\mathcal{H}_{Q_m}|$ elements of $\mathcal{H}_m$, which for sufficiently large $m$ can be bounded as
	\begin{align*}
		r\times |\mathcal{H}_{Q_m}|&\leq m^{|\mathcal{S}|\times |\mathcal{W}|} \times e^{mD_{KL}(p_{\hat{W}|S} Q_{m,S}\|q_{\hat{W}} Q_{m,S})}  \stackrel{(a)}{\leq} e^{m(D_{KL}(p_{\hat{W}|S} Q_{m,S}\|q_{\hat{W}} Q_{m,S})+\nu_1)}\\
		& \stackrel{(b)}{\leq} e^{mE_{Q_{m,S}}[R_S+\nu_1]} \stackrel{(c)}{=} e^{m(\sum_i R_{S_i,W_i}+\nu_1)},
	\end{align*}
	where $(a)$ holds for sufficiently large $m$, $(b)$ by the assumptions of the theorem, and $(c)$ holds when $S^m$ has the empirical distribution $Q_{m,S}$. Now,
	\begin{align*}
		\mathbb{P}_{(S,W)^{\otimes m}} & \left( \min_{j \leq e^{\sum_{i\in [m]}R_{S_i,W_i}+\nu_1}} d(W^m,\hat{\vc{w}}[j];S^m) >\epsilon+\nu_2 \right) \\
		\leq & \mathbb{P}_{(S,W)^{\otimes m}}\left( \min_{j \leq e^{\sum_{i\in [m]}R_{S_i,W_i}+\nu_1}} d(W^m,\hat{\vc{w}}[j];S^m) >\epsilon+\nu_2, \mathcal{T}(S^m,W^m)\in \mathcal{Q}_m\right)\\
		& +\mathbb{P}_{(S,W)^{\otimes m}}\left( \mathcal{T}(S^m,W^m)\notin \mathcal{Q}_m\right)\\
		\stackrel{(a)}{\leq}& P_{(S^m,W^m)}\left( \mathcal{T}(S^m,W^m)\notin \mathcal{Q}_m\right)\\
		\stackrel{(b)}{\leq}& (\delta+\eta_m)^m,	
	\end{align*}
	where $(a)$ holds for sufficiently large values of $m$ and by the hypothesis book construction described above, $(b)$ holds for $\lim_m \eta_m \to 0$ by using \cite[Theorem~11.1.4]{CoverTho06} and since the number of possible empirical distributions of $(s^m,w^m)$ is polynomial with respect to $m$. This completes the proof.

Now, we proceed to show that the condition \eqref{eq:ConditionRenyiSimp} is sufficient. To this end, we use the \emph{random coding} technique, commonly used in information theory \cite{CoverTho06}, for constructing the hypothesis book. For each $\nu_{S,W}$, fix some $p_{\hat{W}|S,\nu_{S,W}} \in \mathcal{Q}(\nu_{S,W})$ and $q_{\hat{W},\nu_{S,W}}$. We show that the algorithm is $(R_{S,W}+\nu_1,\epsilon+\nu_2,\sigma;d_m)$-compressible if \eqref{eq:ConditionRenyiSimp} holds with these choices. The result then follows. For better readability, we drop the dependence on $\nu_{S,W}$, although it is assumed throughout the proof.

Define the set $\mathcal{Q}_m$ as in the first proof. With a slight abuse of notations, we write $\mathcal{T}(s^m)\in \mathcal{Q}_m$, for a sequence $s^m$, if there exists a sequence $w^m$, such that $\mathcal{T}(s^m,w^m)\in \mathcal{Q}_m$. Moreover, if $\mathcal{T}(s^m,w^m)=Q_m$, we write $\mathcal{T}(s^m)=Q_{m,S}$ and $\mathcal{T}(w^m|s^m)=Q_{m,W|S}$.

For each sequence $(s^m,w^m)$, with empirical distribution $Q_m \in \mathcal{Q}_m$, we first construct the random hypothesis book $\mathcal{H}_{m,Q_m}=\{\hat{\vc{W}}_{Q_m}[j],j \in [l_{Q_m}]\}$, where
\begin{align}
	l_{Q_m}\coloneqq e^{m\mathbb{E}_{(S,W)\sim Q_m}[R_{S,W}]+m\nu_1/2}=e^{\sum_{i\in[m]}R_{s_i,w_i}+\nu_1/2}.
\end{align}
For each $\mathcal{H}_{m,Q_m}$, generate in an \iid manner, each $\hat{\vc{W}}_{Q_m}[j]$ according to $q_{\hat{W}}^{\otimes m}$. We construct the \emph{random} $\mathcal{H}_m$ by proper concatenation of  $\mathcal{H}_m \coloneqq \bigcup_{Q_m \in \mathcal{Q}_m} \mathcal{H}_{m,Q_m} \bigcup \{\hat{\vc{w}}_0\}$, and indexing them as explained in the first proof. Consider $m$ is sufficiently large, such that $m^{|\mathcal{S}|\times |\mathcal{W}|}\leq e^{m\nu_1/2}$. Now, in the way that the elements of $\mathcal{H}_m$ are indexed, as explained in the first proof, for each $(s^m,w^m)$ such that $\mathcal{T}(s^m)\in \mathcal{Q}_m$, it is sufficient to consider the first $e^{\sum_{i\in[m]}R_{s_i,w_i}+\nu_1}$ elements.

Note that so far, we have a \emph{random} hypothesis book $\mathcal{H}_m$. In the following, we evaluate the probability of covering failure, \ie LHS of \eqref{eq:condComp}, among all these randomly generated hypothesis books. By showing that this probability is asymptotically less than $\delta+\nu_3$, for any $\nu_3>0$ that is less than $\nu_1/2$, we conclude that there \emph{exists} at least one proper $\mathcal{H}_m$ having such property. The proof completes then by noting that $\nu_3$ can be made arbitrarily small.

Now, denoting the probability among all hypothesis books by the subscript $\mathcal{H}_m$, we have 
\begin{align}
	&\mathbb{P}_{(S,W)^{\otimes m},\mathcal{H}_m} \left( \min_{j \leq e^{\sum_{i\in [m]}R_{S_i,W_i}}} d(W^m,\hat{\vc{W}}[j];S^m) >\epsilon+\nu_2 \right) \nonumber\\
	 &\hspace{0.2 cm}\leq \mathbb{P}_{(S,W)^{\otimes m},\mathcal{H}_m}\left( \min_{j \leq e^{\sum_{i\in [m]}R_{S_i,W_i}+\nu_1}} d(W^m,\hat{\vc{W}}[j];S^m) >\epsilon+\nu_2, \mathcal{T}(S^m,W^m)\in \mathcal{Q}_m\right) \nonumber\\
	& \hspace{0.6 cm} +\mathbb{P}_{(S,W)^{\otimes m},\mathcal{H}_m}\left( \mathcal{T}(S^m,W^m)\notin \mathcal{Q}_m\right) \nonumber\\
	&\hspace{0.2 cm} \stackrel{(a)}{\leq} \mathbb{P}_{(S,W)^{\otimes m},\mathcal{H}_m}\left( \min_{j \leq e^{\sum_{i\in [m]}R_{S_i,W_i}+\nu_1}} d(W^m,\hat{\vc{W}}[j];S^m) >\epsilon+\nu_2, \mathcal{T}(S^m,W^m)\in \mathcal{Q}_m\right) \label{eq:secondProof1}\\
	&\hspace{0.7 cm}+(\delta+\eta_m)^m,	 \nonumber
\end{align}
where $(a)$ follows similarly as the first proof. In the rest of the proof, we show that the first term \eqref{eq:secondProof1}, denoted as $P_1$, can be upper bounded by $e^{-m\nu_3}$ as $m\ra \infty$, for any $\nu_3>0$ that is smaller than $\nu_1/2$. This completes the proof. 

The term $P_1$ can be upper bounded as
\begin{align}
	P_1 \leq \sum_{s_m\colon \mathcal{T}(s^m) \in \mathcal{Q}_m} 	P_{S^{\otimes m}}(s^m)\sum_{\substack{Q_m \in \mathcal{Q}_m}} \mathbb{P}_{(W|S)^{\otimes m}}(\mathcal{T}(s^m,W^m)=Q_m) \times P_2, \label{eq:coveringComp1}
\end{align}
where 
\begin{align}
	P_2\coloneqq \mathbb{P}_{\mathcal{H}_{m},W^m}\left( \min_{j \leq e^{\sum_{i\in [m]}R_{s_i,W_i}+\nu_1}} d(W^m,\hat{\vc{W}}[j];s^m) >\epsilon+\nu_2 \Big| \mathcal{T}(s^m,w^m)\in Q_m\right). \label{eq:coveringComp2}
\end{align}
We continue by analyzing $P_2$ for any $w^m$, such that $\mathcal{T}(s^m,w^m)= Q_m$.

Then,
\begin{align*}
	\mathbb{P}_{\mathcal{H}_{m}}&\left( \min_{j \leq e^{\sum_{i\in [m]}R_{s_i,w_i}+\nu_1}} d(w^m,\hat{\vc{W}}[j];s^m) >\epsilon+\nu_2 \right)\\
	&\leq \mathbb{P}_{\mathcal{H}_{m,Q_m}}\left( \min_{j \leq e^{\sum_{i\in [m]}R_{s_i,w_i}+\nu_1/2}} d(w^m,\hat{\vc{W}}_{Q_m}[j];s^m) >\epsilon+\nu_2 \right)\\
	&\leq \sum_{j\in e^{\sum_{i\in [m]}R_{s_i,w_i}+\nu_1/2}} \mathbb{P}_{\mathcal{H}_{m,Q_m}}\left(d(w^m,\hat{\vc{W}}_{Q_m}[j];s^m) >\epsilon+\nu_2 \right)\\
	&\stackrel{(a)}{\leq} e^{\sum_{i\in [m]}R_{s_i,w_i}+\nu_1/2} e^{-mD_{KL}(p_{\hat{W}|S} Q_{m,S}\|q_{\hat{W}} Q_{m,S})}\\
	&=  e^{m\mathbb{E}_{Q_m}[R_{S,W}+\nu_1/2]} e^{-mD_{KL}(p_{\hat{W}|S} Q_{m,S}\|q_{\hat{W}} Q_{m,S})},	
\end{align*}
where $(a)$ holds by \cite[Theorem~11.1.4]{CoverTho06} and due to the way $\mathcal{H}_{m,Q_m}$ is constructed and since $p_{\hat{W}|S}$ satisfies the distortion criterion \eqref{eq:tailDistortionSimp}.

Now, Combining this with \eqref{eq:coveringComp1} and \eqref{eq:coveringComp2}, and since 
\begin{align*}
	\mathbb{P}_{(W|S)^{\otimes m}}(\mathcal{T}(s^m,W^m)=Q_m) \leq  e^{-mD_{KL}(Q_{m}\|P_{W|S}Q_{m,S})},
\end{align*}
by \cite[Theorem~11.1.4]{CoverTho06}, we conclude that 
\begin{align}
	P_1 \leq& m^{|\mathcal{S}|\times |\mathcal{W}|}  e^{m\mathbb{E}_{Q_m}[R_{S,W}+\nu_1/2]} e^{-mD_{KL}(p_{\hat{W}|S} Q_{m,S}\|q_{\hat{W}} Q_{m,S})}  e^{-mD_{KL}(Q_{m}\|P_{W|S}Q_{m,S})} \nonumber\\
	\leq & e^{-m\nu_3}, \label{eq:coveringComp3}
\end{align}
where the last step holds for any $\nu_1/2>\nu_3>0$, given that $m$ is sufficiently large, due to Condition~\ref{eq:condComp}.


 \subsection{Proof of Theorem~\ref{th:generalTail}} \label{pr:generalTail}
\subsubsection{Part i.}
We start the proof by showing a potentially-data-dependent version of the \cite[Lemma~24]{Sefidgaran2022},
tailored for the stochastic learning setup. The lemma is proved in Appendix~\ref{pr:tailKL}.

\begin{lemma} \label{lem:tailKL} For any $\delta>0$,
		\begin{align}
			&\log \mathbb{P}\left(f(S,W) > \Delta(S,W) \right)\leq \max \bigg(\log(\delta), \label{eq:tailKL}\\
			&\hspace{-0.4 cm}\sup_{\nu_{S,W}\in \mathcal{F}_{S,W}^{\delta}} \inf_{p_{\hat{W}|S}\in \mathcal{Q}(\nu_{S,W})}\inf_{\lambda \geq 0}\left\{-D_{KL}(\nu_{S,W}\|P_{S,W})-\lambda \left(\mathbb{E}_{\nu_{S,W}}[\Delta(S,W)]-\epsilon-\mathbb{E}_{\nu_S P_{\hat{W}|S}} \left[g(S,\hat{W})\right] \right)\right\}\bigg). \nonumber
		\end{align} 
	\end{lemma}
	
	In the rest, we show that for any $\nu_{S,W}$, $p_{\hat{W}|S}$, $\lambda \geq 0$, and $q_{\hat{W}|S}$, we have 
	\begin{align}
		-D_{KL}(\nu_{S,W}\|P_{S,W})+\lambda \mathbb{E}_{\nu_S P_{\hat{W}|S}} \left[g(S,\hat{W})\right] \leq  & D_{KL}\left(p_{\hat{W}|S} \nu_{S}\|q_{\hat{W}|S} \nu_{S}\right) +\log\mathbb{E}_{P_S q_{\hat{W}|S}}\left[e^{\lambda g(S,\hat{W})}\right]\nonumber \\
		& -D_{KL}(\nu_{S,W}\|P_{W|S} \nu_{S}). \label{eq:tailBound1}
	\end{align}
	This inequality and Lemma~\ref{lem:tailKL} complete the proof of the theorem. 	To show this inequality, for any $S=s$,  using the Donsker-Varadhan inequality, we have 
	\begin{align*}
		\lambda \mathbb{E}_{P_{\hat{W}|S=s}} \left[g(s,\hat{W})\right] \leq D_{KL}\left(P_{\hat{W}|S=s}\|q_{\hat{W}|S=s}\right)+\log\mathbb{E}_{q_{\hat{W}|S=s}}\left[e^{\lambda g(s,\hat{W})}\right].
	\end{align*}
	Next, taking the expectation with respect to $\nu_S$, gives
	\begin{align}
		\lambda \mathbb{E}_{\nu_S P_{\hat{W}|S}} \left[g(S,\hat{W})\right] &\leq  D_{KL}\left(P_{\hat{W}|S}\nu_S\|q_{\hat{W}|S}\nu_S\right)+\mathbb{E}_{\nu_S}\log\left(\mathbb{E}_{q_{\hat{W}|S}}\left[e^{\lambda g(S,\hat{W})}\right]\right) \nonumber\\
		&\stackrel{(a)}{\leq}  D_{KL}\left(P_{\hat{W}|S}\nu_S\|q_{\hat{W}|S}\nu_S\right)+D_{KL}(\nu_S\|P_S)+\log\mathbb{E}_{P_S}\left[\mathbb{E}_{q_{\hat{W}|S}}\left[e^{\lambda g(S,\hat{W})}\right]\right],\label{eq:tailBound2}
	\end{align}
	where $(a)$ is derived by using the Donsker-Varadhan inequality. The desired inequality~\eqref{eq:tailBound1} can be derived from \eqref{eq:tailBound2}. This completes the proof.

 	\subsubsection{Part ii.} To show this part, we first establish a lemma similar to Lemma~\ref{lem:tailKL}, proved in Appendix~\ref{pr:tailRenyi}.
	
	\begin{lemma} \label{lem:tailRenyi} For any $\alpha>1$ and $\delta>0$,
		\begin{align}
			&\log \mathbb{P}\left(f(S,W) > \Delta(S,W) \right)\leq \max \Bigg(\log(\delta), \label{eq:tailRenyi}\\
			&\sup_{\nu_{S,W}\in \mathcal{F}_{S,W}^{\delta}} \inf_{\lambda \geq 0}\left\{-D_{KL}(\nu_{S,W}\|P_{S,W})-\lambda \mathbb{E}_{\nu_{S}}\left[\log \mathbb{E}_{\nu_{W|S}}[\Delta(S,W)]-\log\mathbb{E}_{\nu_{W|S}}\left[f(S,W)\right] \right]\right\}\Bigg). \nonumber
		\end{align} 
	\end{lemma}
	Now,  for $\lambda \geq 1/C_{\alpha}$ where $C_{\alpha}\coloneqq (\alpha-1)/\alpha$, we have
	\begin{align*}
		\lambda \mathbb{E}_{\nu_{S}}\left[\log\mathbb{E}_{\nu_{W|S}}\left[f(S,W)\right] \right]=& \frac{\lambda C_{\alpha} }{C_{\alpha} }\mathbb{E}_{\nu_{S}}\left[\log\mathbb{E}_{\nu_{W|S}}\left[f(S,W)\right] \right]\\
		\stackrel{(a)}{\leq}&   \frac{1}{C_{\alpha}}\mathbb{E}_{\nu_{S}}\left[\log\mathbb{E}_{\nu_{W|S}}\left[f(S,W)^{\lambda C_{\alpha} }\right] \right]\\
		\stackrel{(b)}{\leq} &  \frac{1}{C_{\alpha}}\mathbb{E}_{\nu_{S}}\left[C_{\alpha} D_{\alpha} \left(\nu_{W|S}\|q_{W|S}\right)+ C_{\alpha} \log\mathbb{E}_{q_{W|S}}\left[f(S,W)^{\frac{\lambda C_{\alpha}  }{C_{\alpha}}}\right]\right]\\
		\leq & \mathbb{E}_{\nu_{S}}\left[D_{\alpha} \left(\nu_{W|S}\|q_{W|S}\right)\right]+ D_{KL}(\nu_S\|P_S)+  \log\mathbb{E}_{P_Sq_{W|S}}\left[f(S,W)^{\lambda }\right]\\
		= & \mathbb{E}_{\nu_{S}}\left[D_{\alpha} \left(\nu_{W|S}\|q_{W|S}\right)\right]+ D_{KL}(\nu_{S,W}\|P_{S,W})\\
		&-D_{KL}(\nu_{S,W}\|P_{W|S}\nu_S) +  \log\mathbb{E}_{P_Sq_{W|S}}\left[f(S,W)^{\lambda }\right],
	\end{align*}
	where $(a)$ is due to convexity of the function $x^{r}$ for $r \geq 1$, and $(b)$ by the generalization of the Donsker-Varadhan inequality \cite{atar2015robust}. The proof completes by using Lemma~\ref{lem:tailRenyi}.


 \subsection{Proof of Theorem~\ref{th:generalTailExp}} \label{pr:generalTailExp}
Denote
	\begin{align*}
		\mathcal{A}_{S} \coloneqq \Big\{s\in \mathcal{S}\,\big|\, \exists \pi \colon \mathbb{E}_{W\sim \pi}\left[f(s,W)\right]> \Delta(s,\pi) \Big\}.
	\end{align*}
	Consider a learning algorithm $P_{W|S}$ defined as follows: if $s \in \mathcal{A}_S$, then let $P_{W|S=s}$ be any distribution $\pi_{s}$ such that $\mathbb{E}_{W\sim \pi_s}\left[f(s,W)\right]> \Delta(s,\pi_s)$, otherwise choose $P_{W|S=s}$ as any arbitrary distribution. Then, we have
	\begin{align*}
		\mathbb{P}_S\left(	\exists \pi: \mathbb{E}_{W\sim \pi}\left[f(S,W)\right]> \Delta(S,\pi)\right)=	\mathbb{P}_S\left(\mathbb{E}_{W\sim P_{W|S}}\left[f(S,W)\right]> \Delta(S,P_{W|S})\right).
	\end{align*}
	Hence, it is sufficient to show the RHS is less than $\delta$, given that (by assumptions of the theorem) any of these conditions hold:
	\begin{itemize}[leftmargin=*]
		\item[i.] For any distribution $\nu_{S} \in \mathcal{F}_{S}^{\delta} \coloneqq \mathcal{G}_{S}^{\delta}\bigcap\mathcal{S}_{S}\left( \mathbb{E}_{W \sim P_{W|S}}[f(s,W)]- \Delta(s,P_{W|S})\right)$,
		\begin{align}
			\log(\delta) \geq \inf_{p_{\hat{W}|S}\in \mathcal{Q}( \nu_{S} P_{W|S})}\inf_{q_{\hat{W}|S},\lambda > 0}  \bigg\{&		\mathfrak{T}_{1,P_S}(\nu_S,p_{\hat{W}|S},q_{\hat{W}|S},\lambda g)-\lambda \left( \mathbb{E}_{\nu_{S}}\left[\Delta(S,P_{W|S})\right]-\epsilon\right) \bigg\}, \label{eq:ConditionTailExp2}
		\end{align}
		where  $\mathcal{Q}(\nu_{S}P_{W|S})$ contains all the distributions $p_{\hat{W}|S}$ such that 
		\begin{align*}
			\mathbb{E}_{\nu_{S}p_{\hat{W}|S}}\left[\Delta(S,P_{W|S})-g(S,\hat{W})\right] \leq \epsilon.
		\end{align*}  
		\item[ii.] For $\alpha>1$ and any distribution $\nu_S \in \mathcal{F}_S^{\delta}$,
		\begin{align*}
			\inf_{q_{W|S},\lambda \geq \alpha/(\alpha-1)}  \bigg\{&\mathfrak{T}_{\alpha,P_S}(\nu_{S},P_{W|S},q_{W|S},\lambda\log(f) -\lambda \mathbb{E}_{\nu_{S}}\left[\log \Delta(S,P_{W|S})\right]\bigg\} \leq \log(\delta),
		\end{align*}
	\end{itemize}
	We show the proof of each case separately.
	
	\begin{itemize}[leftmargin=*]
		\item[i.] Applying Lemma~\ref{lem:tailKL} with $f(S,W) \mapsto \mathbb{E}_{P_{W|S}}\left[f(S,W)\right]$ and  $\Delta(S,W) \mapsto \Delta(S,P_{W|S})$ (note that both terms depend only on $S$) gives
		\begin{align}
			&\log \mathbb{P}\left(\mathbb{E}_{P_{W|S}}\left[f(S,W)\right] \geq \Delta(S,P_{W|S}) \right)\leq \max \Bigg(\log(\delta), \nonumber\\
			&\sup_{\nu_{S}\in \mathcal{F}_{S}^{\delta}} \inf_{p_{\hat{W}|S}\in \mathcal{Q}(\nu_{S}P_{W|S})}\inf_{\lambda \geq 0}\left\{-D_{KL}(\nu_{S}\|P_{S})-\lambda \left(\mathbb{E}_{\nu_{S}}[\Delta(S,P_{W|S})]-\epsilon-\mathbb{E}_{\nu_S p_{\hat{W}|S}} \left[g(S,\hat{W})\right] \right)\right\}\Bigg). \nonumber
		\end{align} 
		The rest of the proof follows similarly to the proof of part i. of Theorem~\ref{th:generalTail} to upper bound the term $\lambda \mathbb{E}_{\nu_S p_{\hat{W}|S}} \left[g(S,\hat{W})\right]$.
		\item[ii.] Applying Lemma~\ref{lem:tailRenyi} with $f(S,W) \mapsto \mathbb{E}_{P_{W|S}}\left[f(S,W)\right]$ and  $\Delta(S,W) \mapsto \Delta(S,P_{W|S})$ (note that both terms depend only on $S$) gives
		\begin{align}
			\log \mathbb{P}\Big(\mathbb{E}_{P_{W|S}}&\left[f(S,W)\right] \geq \Delta(S,P_{W|S}) \Big)\leq \max \Bigg(\log(\delta), \nonumber\\
			&\sup_{\nu_{S}\in \mathcal{F}_{S}^{\delta}} \inf_{\lambda \geq 0}\left\{-D_{KL}(\nu_{S}\|P_{S})-\lambda \mathbb{E}_{\nu_{S}}\left[\log \Delta(S,P_{W|S})-\log\mathbb{E}_{P_{W|S}}\left[f(S,W)\right] \right]\right\}\Bigg). \nonumber
		\end{align} 
		The rest of the proof follows similarly to the proof of part ii. of Theorem~\ref{th:generalTail} to bound  $\lambda \mathbb{E}_{\nu_{S}}\left[\log\mathbb{E}_{P_{W|S}}\left[f(S,W)\right] \right]$.
	\end{itemize}


 \subsection{Proof of Theorem~\ref{th:expectation}} \label{pr:expectation}

 \subsubsection{Part i.}	
First, note that
	\begin{align}
		\mathbb{E}_{(S,W)\sim P_{S,W}}\left[f(S,W)\right] \leq 	\mathbb{E}_{(S,\hat{W})\sim P_{S}p_{\hat{W}|S}}\left[g(S,\hat{W})\right]+\epsilon. \label{eq:exp1}
	\end{align}	
	Now, using the Donsker-Varadhan inequality, for any $p_{\hat{W}|S}$, $\lambda \geq 0$, and $q_{\hat{W}|S}$, we have 
	\begin{align}
		\lambda \mathbb{E}_{P_S p_{\hat{W}|S}} \left[g(S,\hat{W})\right] \leq D_{KL}\left(p_{\hat{W}|S} P_{S}\|q_{\hat{W}|S} P_{S}\right) +\log\mathbb{E}_{P_S q_{\hat{W}|S}}\left[e^{\lambda g(S,\hat{W})}\right]. \label{eq:expBound1}
	\end{align}
	Combining this inequality and inequality~\eqref{eq:exp1} completes the proof.
 \subsubsection{Part ii.}
Using the generalization of the Donsker-Varadhan inequality \cite{atar2015robust}, for any $\lambda \geq 1/C_{\alpha}$, where $C_{\alpha}\coloneqq (\alpha-1)/\alpha$, and $q_{W|S}$, we have 
	\begin{align*}
		\lambda \log\mathbb{E}_{P_{S,W}}\left[f(S,W)\right] =& \frac{\lambda C_{\alpha} }{C_{\alpha} }\log\mathbb{E}_{P_{S,W}}\left[f(S,W)\right]\\
		\leq&   \frac{1}{C_{\alpha}}\log\mathbb{E}_{P_{S,W}}\left[f(S,W)^{\lambda C_{\alpha} }\right]\\
		\leq&  \frac{1}{C_{\alpha}}\left[C_{\alpha} D_{\alpha} \left(P_{S,W}\|q_{W|S}P_S\right)+ C_{\alpha} \log\mathbb{E}_{P_Sq_{W|S}}\left[f(S,W)^{\frac{\lambda C_{\alpha}  }{C_{\alpha}}}\right]\right].
	\end{align*}
	This completes the proof.


 \subsection{Proof of Proposition~\ref{prop:PAClossy}} \label{pr:PAClossy}
The proof of part i. follows immediately from part i. of Theorem~\ref{th:generalTailExp}. To show the second part, we use part i. of Theorem~\ref{th:generalTail}.  For each $\nu_{S,W} \in \mathcal{F}_{S,W}^{\delta}$, consider $p_{\hat{W}|S}$ as the induced conditional distribution under $\nu_{S,W}p_{\hat{W}|S,W}$, where $p_{\hat{W}|S,W}$ satisfies $p_{\hat{W}|S,W}=p_{\hat{W}|W}$. Let $\lambda=1$ and $\Delta(S,W)$ be the right-hand side of part ii. Then, we have
	\begin{align}
	&\mathfrak{T}_{1,P_{S}}(\nu_{S},p_{\hat{W}|S},q_{\hat{W}|S},\lambda g) {-}D_{KL}(\nu_{S,W} \|P_{W|S}\nu_S){-}\lambda \left( \mathbb{E}_{\nu_{S,W}}\left[\Delta(S,W)\right]-\epsilon\right) \nonumber \\
		&\hspace{0.1 cm} \leq \mathbb{E}_{\nu_S}\left[D_{KL}\left(p_{\hat{W}|S}\|q_{\hat{W}|S}\right)\right]-D_{KL}(\nu_{S,W} \|P_{W|S}\nu_S)- \mathbb{E}_{\nu_{S,W}p_{\hat{W}|W}}\left[\log\left(\frac{\der p^*_{\hat{W}|S}} {\der q_{\hat{W}|S}}\right)\right]{+}\log(\delta)\nonumber \\
		&\hspace{0.1 cm} \stackrel{(a)}{\leq} \mathbb{E}_{\nu_S}\left[D_{KL}\left(p_{\hat{W}|S}\|q_{\hat{W}|S}\right)-D_{KL}(p_{\hat{W}|S} \|p^*_{\hat{W}|S})\right]- \mathbb{E}_{\nu_{S,W}p_{\hat{W}|W}}\left[\log\left(\frac{\der p^*_{\hat{W}|S}} {\der q_{\hat{W}|S}}\right)\right]+\log(\delta) \nonumber \\
		&\hspace{0.1 cm}= \mathbb{E}_{\nu_Sp_{\hat{W}|S}}\left[\log\left(\frac{\der p^*_{\hat{W}|S}} {\der q_{\hat{W}|S}}\right)\right]- \mathbb{E}_{\nu_{S,W}p_{\hat{W}|W}}\left[\log\left(\frac{\der p^*_{\hat{W}|S}} {\der q_{\hat{W}|S}}\right)\right]+\log(\delta) \nonumber \\
		&\hspace{0.1 cm}=\log(\delta),
	\end{align}
	where $(a)$ is deduced from the data processing inequality. This completes the proof.


 \subsection{Proof of Theorem~\ref{th:RDProcessSup}} \label{pr:RDProcessSup} 
We use Theorem~\ref{th:generalTailExp} to prove the result. Let
	\begin{align*}
		f(s,w^{\T}) &\coloneqq \gen(S,W^{\T})=\frac{1}{\dT}\sum_{t\in \T} \gen(s,w^t), \\
		\Delta(\pi) &\coloneqq  \sqrt{\frac{\sup_{\nu_s \in \mathcal{G}^{\delta}_S}\mathfrak{RD}(\nu_S,\epsilon;\pi)+\log(1/\delta)}{2n}}+\epsilon.
	\end{align*}
	Let $\mathcal{\hat{W}}\subseteq \mathcal{W}$ and $g(s,\hat{w}^{\T})=f(s,\hat{w}^{\T})$. Now, consider any arbitrary distributions $\pi_S$, index by $S$, and any $\nu_S \in \mathcal{F}_{S}^{\delta}$. Consider any $p_{\hat{W}^{\T}|S}$ such that 
	\begin{align*}
		\mathbb{E}_{\nu_{S} \pi_S p_{\hat{W}^{\T}|S}}\left[f(S,W^{\T})-f(S,\hat{W}^{\T})\right] \leq \epsilon.
	\end{align*}
	Since $\nu_S \in \mathcal{F}_{S}^{\delta}$, then $p_{\hat{W}^{\T}|S}$ satisfies the distortion criterion \eqref{eq:distExp}. Let $q_{\hat{W}^{\T}|S}\coloneqq q_{\hat{W}^{\T}}$ be the marginal distribution of $\hat{W}^{\T}$ under $p_{\hat{W}^{\T}|S} \nu_S$. Then, LHS of Condition~\ref{eq:ConditionTailExp} becomes as 
	\begin{align}
		\inf_{p_{\hat{W}^{\T}|S},q_{\hat{W}^{\T}},\lambda>0}\bigg\{I_{\nu_s p_{\hat{W}^{\T}|S}}(S;\hat{W}^{\T})-\lambda \big( \mathbb{E}_{\nu_{S}}&\left[\Delta(\pi_S)\right]-\epsilon\big)+\log\mathbb{E}_{P_S q_{\hat{W}^{\T}}}\left[e^{\lambda f(S,\hat{W}^{\T})}\right] \bigg\} \nonumber \\
		\stackrel{(a)}{\leq}& \inf_{\lambda > 0}  \bigg\{\mathfrak{RD}(\nu_S,\epsilon;\pi_S) -\lambda \left(\Delta(\pi_S)-\epsilon\right)+\frac{\lambda^2 }{8n} \bigg\} \nonumber \\
		\stackrel{(b)}{=}& \mathfrak{RD}(\nu_S,\epsilon;\pi_S) - \sup_{\nu_s \in \mathcal{G}^{\delta}_S}\mathfrak{RD}(\nu_S,\epsilon;\pi_S)-\log(1/\delta)\nonumber \\
		\leq & \log(\delta)
	\end{align}
	where $(a)$ is concluded since $f(S,\hat{w}^{\T})$ is $1 /(2\sqrt{n})$-subgaussian and $(b)$ is derived for $\lambda \coloneqq 4n(\Delta(\pi_S)-\epsilon)$. This completes the proof of part i.


 \subsection{Proof of Theorem~\ref{th:RDProcess}} \label{pr:RDProcess} 
	We use Theorem~\ref{th:generalTailExp} with $\epsilon \mapsto 4 \mathfrak{L}\epsilon$ to prove the result. Let 
	\begin{align*}
		f(s,w^{\T}) &\coloneqq \gen(S,W^{\T})^2=\left(\frac{1}{\dT}\sum_{t\in \T} \gen(s,w^t)\right)^2, \\
		\Delta(s,\pi) &\coloneqq  \frac{\mathfrak{RD}(s,\epsilon;\pi)+\log(\sqrt{2n}M/\delta)}{2n-1}+4\mathfrak{L}\epsilon.
	\end{align*}
	Let $\mathcal{\hat{W}}\subseteq \mathcal{W}$ and $g(s,\hat{w}^{\T})=f(s,\hat{w}^{\T})$. Theorem~\ref{th:generalTailExp} remains valid, if instead of all $\pi$, we limit ourselves to only one distribution $P_{W^{\T}|S}$, that we denote for the ease of exposition as $\pi_S$.  Consider any $\nu_S \in \mathcal{F}_{S}^{\delta}$. 	For any $s \in \supp(\nu_S)$, first let $p_{\hat{W}^{\T}|W^{\T},s}$ be any distribution that satisfies
	\begin{align*}
		\mathbb{E}_{\pi_s p_{\hat{W}^{\T}|W^{\T},s}}\left[\rho(W^{\T},\hat{W}^{\T})\right] \leq \epsilon.
	\end{align*}
	In the following, for simplicity, we denote by $Q$, any (possibly marginal) distribution induced by $\nu_S \pi_S p_{\hat{W}^{\T}|W^{\T},S}$, \eg $Q_{W^{\T},\hat{W}^{\T}}$ is the marginal distribution of $(W^{\T},\hat{W}^{\T})$ under the joint mentioned distribution. Note that
	\begin{align}
		\mathbb{E}_{Q_{W^{\T},\hat{W}^{\T}}}\left[ \rho(W^{\T},\hat{W}^{\T})\right]=& \mathbb{E}_{\nu_S} \mathbb{E}_{\pi_s p_{\hat{W}^{\T}|W^{\T},s}}\left[\rho(W^{\T},\hat{W}^{\T})\right] \leq  \epsilon. \label{eq:rdpr1}
	\end{align}
	Now define another distribution, denoted by $\tilde{Q}$, as any (possibly marginal) distribution induced by $\nu_S \pi_S \tilde{p}_{\hat{W}^{\T}|W^{\T},S}$, where $\tilde{p}_{\hat{W}^{\T}|W^{\T},S}\coloneqq Q_{\hat{W}^{\T}|W^{\T}}$. Note that $S\markov W^{\T}\markov \hat{W}^{\T}$ forms a Markov chain under $\tilde{Q}$, \ie $\tilde{Q}_{\hat{W}^{\T}|W^{\T},S}=\tilde{Q}_{\hat{W}^{\T}|W^{\T}}$. Furthermore, $\tilde{Q}_{W^{\T},\hat{W}^{\T}}\coloneqq Q_{W^{\T},\hat{W}^{\T}}$.
	
	Now,
	\begin{align*}
		\mathbb{E}_{\nu_S  \tilde{Q}_{\hat{W}^{\T}|S}}\bigg[\Delta(S,\pi_S)-\gen(S,\hat{W}^{\T})^2\bigg]\nonumber 
		\stackrel{(a)}{\leq}& \mathbb{E}_{\nu_S  \tilde{Q}_{\hat{W}^{\T}|S}}\left[\mathbb{E}_{\pi_S}\left[\gen(S,W^{\T})^2\right]-\gen(S,\hat{W}^{\T})^2\right]\nonumber \\
		=&	\mathbb{E}_{\nu_S \pi_S \tilde{Q}_{\hat{W}^{\T}|S,W^{\T}}}\left[\gen(S,W^{\T})^2-\gen(S,\hat{W}^{\T})^2\right]\nonumber \\
		= &\mathbb{E}_{\tilde{Q}_{S,W^{\T},\hat{W}^{\T}}} \left[\gen(S,W^{\T})^2-\gen(S,\hat{W}^{\T})^2\right]\nonumber \\
		\leq &\mathbb{E}_{\tilde{Q}_{S,W^{\T},\hat{W}^{\T}}} \left[\big|\gen(S,W^{\T})^2-\gen(S,\hat{W}^{\T})^2\big|\right]\nonumber \\
		\stackrel{(b)}{\leq} &2\mathbb{E}_{\tilde{Q}_{S,W^{\T},\hat{W}^{\T}}} \left[\big|\gen(S,W^{\T})-\gen(S,\hat{W}^{\T})\big|\right]\nonumber \\
		\stackrel{(c)}{\leq}&4\mathfrak{L}\,\mathbb{E}_{\tilde{Q}_{S,W^{\T},\hat{W}^{\T}}} \left[\rho(W^{\T},\hat{W}^{\T})\right]\nonumber \\
		= &4\mathfrak{L}\,\mathbb{E}_{\tilde{Q}_{W^{\T},\hat{W}^{\T}}} \left[\rho(W^{\T},\hat{W}^{\T})\right]\nonumber \\
		\stackrel{(d)}{=} &4\mathfrak{L}\,\mathbb{E}_{Q_{W^{\T},\hat{W}^{\T}}} \left[\rho(W^{\T},\hat{W}^{\T})\right]\nonumber \\
		\stackrel{(e)}{\leq} & 4\mathfrak{L} \epsilon,
	\end{align*}
	where $(a)$ holds since $\nu_S \in \mathcal{F}_{S}^{\delta}$, $(b)$ due to the fact that $\ell(z,w)\in [0,1]$, $(c)$ by the Lipschitz assumption, $(d)$ is derived by noting that $Q_{W^{\T},\hat{W}^{\T}}=\tilde{Q}_{W^{\T},\hat{W}^{\T}}$, and $(e)$ due to \eqref{eq:rdpr1}.
	
	Hence, $p_{\hat{W}^{\T}|S} \coloneqq \tilde{Q}_{\hat{W}^{\T}|S}$ is in the set $\mathcal{Q}(\nu_S\pi_S)$, as defined in Theorem~\ref{th:generalTailExp}. Let $q_{\hat{W}^{\T}|S}\coloneqq q_{\hat{W}^{\T}}\coloneqq\tilde{Q}_{\hat{W}^{\T}}$. Note that $D_{KL}\left(p_{\hat{W}^{\T}|S}\nu_S\|q_{\hat{W}^{\T}|S}\nu_S\right)=I_{\tilde{Q}}(S;\hat{W}^{\T})$.	
		
    Now, restricting the inf in LHS of Condition~\ref{eq:ConditionTailExp} to only such choices described above, the LHS can be upper bounded as
		\begin{align}
			\text{LHS}&\leq \inf_{p_{\hat{W}^{\T}|S},q_{\hat{W}^{\T}}\lambda>0}\bigg\{I_{\tilde{Q}}(S;\hat{W}^{\T})-\lambda \left( \mathbb{E}_{\nu_{S}}\left[\Delta(S,\pi_S)\right]-4\mathfrak{L}\epsilon\right)+\log\mathbb{E}_{P_S q_{\hat{W}^{\T}}}\left[e^{\lambda f(S,\hat{W}^{\T})}\right] \bigg\} \nonumber \\
			&\stackrel{(a)}{\leq} \inf_{p_{\hat{W}^{\T}|S},q_{\hat{W}^{\T}}}\bigg\{I_{\tilde{Q}}(S;\hat{W}^{\T})-(2n-1) \left( \mathbb{E}_{\nu_{S}}\left[\Delta(S,\pi_S)\right]-4\mathfrak{L}\epsilon\right)+\log(\sqrt{2n})\bigg\} \nonumber \\
			&\stackrel{(b)}{\leq} \inf_{p_{\hat{W}^{\T}|S},q_{\hat{W}^{\T}}}\bigg\{I_{\tilde{Q}}(W^{\T};\hat{W}^{\T})-(2n-1) \left( \mathbb{E}_{\nu_{S}}\left[\Delta(S,\pi_S)\right]-4\mathfrak{L}\epsilon\right)+\log(\sqrt{2n})\bigg\} \nonumber \\
			&\stackrel{(c)}{=} \inf_{p_{\hat{W}^{\T}|S},q_{\hat{W}^{\T}}}\bigg\{I_{Q}(W^{\T};\hat{W}^{\T})-(2n-1) \left( \mathbb{E}_{\nu_{S}}\left[\Delta(S,\pi_S)\right]-4\mathfrak{L}\epsilon\right)+\log(\sqrt{2n})\bigg\} \nonumber \\
			&\stackrel{(d)}{\leq} \inf_{p_{\hat{W}^{\T}|S},q_{\hat{W}^{\T}}}\bigg\{I_{Q}(W^{\T};S,\hat{W}^{\T})-(2n-1) \left( \mathbb{E}_{\nu_{S}}\left[\Delta(S,\pi_S)\right]-4\mathfrak{L}\epsilon\right)+\log(\sqrt{2n})\bigg\} \nonumber \\
			&\stackrel{(e)}{=} \inf_{p_{\hat{W}^{\T}|S},q_{\hat{W}^{\T}}}\bigg\{I_{Q}(W^{\T};\hat{W}^{\T}|S)+I_Q(W^{\T};S)-(2n-1) \left( \mathbb{E}_{\nu_{S}}\left[\Delta(S,\pi_S)\right]-4\mathfrak{L}\epsilon\right)+\log(\sqrt{2n})\bigg\} \nonumber \\
			&\leq \mathbb{E}_{\nu_S}\left[\mathfrak{RD}(S=s,\epsilon;\pi_{S=s})\right]+I_Q(W^{\T};S)-(2n-1) \left( \mathbb{E}_{\nu_{S}}\left[\Delta(S,\pi_S)\right]-4\mathfrak{L}\epsilon\right)+\log(\sqrt{2n}) \nonumber \\
			&= \log(\delta)+\mathbb{E}_{\nu_S,\pi_S} \left[\log \frac{\der\pi_S}{\der [\pi_S \nu_S]_{W^{\T}}} \right]
			-\mathbb{E}_{\nu_S} \left[\sup_{\nu \in \mathcal{G}_{S}^{\delta}} \mathbb{E}_{\pi_S}  \log \frac{\der\pi_S}{\der [\pi_S \nu]_{W^{\T}}} \right]\nonumber \\
			&\leq \log (\delta),
		\end{align}
		where $\der [\pi_S \nu]_{W^{\T}}$ (similarly for $\nu_S$) denotes the marginal distribution of $W^{\T}$ under $\pi_S \nu$, 
		\begin{itemize}
			\item[$(a)$] holds for $\lambda\coloneqq 2n-1$ and since for each $\hat{w}^{\T}$, $\gen(S,\hat{w}^{\T})$ is $1/(2\sqrt{n})$-subgaussian, and hence $\mathbb{E}\left[e^{(2n-1)f(S,\hat{w}^{\T})}\right] \leq \sqrt{2n}$ due to \cite[Theorem~2.6.IV.]{wainwright_2019},
			\item[$(b)$] is due to the data processing inequality by using the fact that under $\tilde{Q}$,  $S\markov W^{\T}\markov \hat{W}^{\T}$ forms a Markov chain,
			\item[$(c)$] is deduced, since the marginal distributions of $(W^{\T},\hat{W}^{\T})$ is the same under $Q$ and $\tilde{Q}$ by definition,
			\item[$(d)$] holds by the data processing inequality,
			\item[$(e)$] is derived by using the chain rule for mutual information.
		\end{itemize}
		This completes the proof.

\subsection{Proofs of Equations~\texorpdfstring{\eqref{in-expectation-bound-generalization-error-counter-example}}{(38)} and~\texorpdfstring{\eqref{tail-bound-generalization-error-counter-example}}{(39)}} \label{pr:example}
\subsubsection{Proof of~\texorpdfstring{\eqref{in-expectation-bound-generalization-error-counter-example}}{(38)}}

We start by recalling some notations and lemmas defined and developed in \cite{haghifam2023limitations}. Define $\vc{B}=(B_1,B_2,\ldots,B_d) \in \{0,1\}^d$ as following: for $j\in[d]$, $B_j=1$ if and only if for all $i\in[n]$, $Z_{i,j}=0$. Otherwise, \ie if at least for one $i\in[n]$, $Z_{i,j}=1$, $B_j=0$. Trivially, $B_j=1$ with probability $2^{-n}$. When $B_j=1$, we call that coordinate, a `bad' coordinate. Let $\|\vc{B}\|_0 =\sum_{j\in[d]} B_j$ denote the number of bad coordinates. Denote the ordered indices of the bad coordinates as $\mathcal{B}=\{i_1,\ldots,i_{\|\vc{B}\|_0}\}\subseteq [d]$. As shown in \cite[Equation~(15)]{haghifam2023limitations}, we have
\begin{align}
    \mathbb{P}\left(T/2\leq \|\vc{B}\|_0 \leq T  \right) \geq 1-2e^{-T/36}. \label{eq:badCnumbers}
\end{align}

By denoting $\hat{\mu}_j$ as the mean of coordinate $j$ of the samples of $S$, \ie letting $\hat{\mu}_j \coloneqq \frac{1}{n}\sum_{i\in[n]} Z_{i,j}$, we can re-write the empirical risk for any $\vc{w}=(w_1,w_2,\ldots,w_d)\in \mathcal{W}$ as
\begin{align}
    \hat{\mathcal{L}}(S,\vc{W}) = \sum_{j\in[d]} \hat{\mu}_j w_j^2 + \lambda \langle \hat{\mu},\vc{w}\rangle + \max \left\{\max_{j\in[d]}\{w_j\},0\right\},
\end{align}
where  $\hat{\mu}=(\hat{\mu}_1,\hat{\mu}_2,\ldots,\hat{\mu}_d)$.

Next, we state a lemma on the dynamics of GD, developed in \cite{haghifam2023limitations}, using the former results of \cite{bassily2020stability,amir2021sgd}. 

\begin{lemma}[{ \cite[Lemma~18]{haghifam2023limitations}}] \label{lem:GDdynamics} Under the event that $\{T/2\leq \|\vc{B}\|_0 \leq T \}$,
\begin{align}
    \vc{W}_j^{\{T\}} = \begin{cases} \frac{\lambda}{2}\left(-1 + \left(1-2\eta \hat{\mu}_j\right)^T\right), & \text{if } j\in [d]\setminus \mathcal{B}, \\ -\eta, & \text{if } j \in \mathcal{B}.
    \end{cases}
\end{align}
\end{lemma}
Let $\mathcal{E}$ be a binary random variable which is equal to $0$ whenever $\{T/2\leq \|\vc{B}\|_0 \leq T\}$, and equal to $1$, otherwise. Denote
\begin{align*}
 v_0=&\frac{\lambda}{2}\left(-1+(1-\eta)^T\right),\\
 v_1=&-\eta.
\end{align*}
 Fix a given $r\in[1-\frac{1}{n^2},1]$ and consider the following quantization law of $p_{\hat{\vc{W}}|S}$: if $\mathcal{E}=1$, then let
 \begin{align}
     \hat{\vc{W}}=(\hat{W}_1,\ldots,\hat{W}_d)=\vc{0}\in \mathbb{R}^d,
 \end{align} with probability one. Otherwise, \ie if $\mathcal{E}=0$, then for each $j\in[d]$, if $\hat{\mu}_j=0$, then let
\begin{align}
    \hat{W}_j = \begin{cases} v_0, & \text{with probability $r$}, \\
    v_1, & \text{with probability $1-r$},    
    \end{cases}
\end{align}
and if $\hat{\mu}_j>0$, then let $\hat{W}_j = v_0$ with probability one.

It can be easily verified that for every $\hat{\vc{w}}$, with positive probability under $P_S p_{\hat{\vc{W}}|S}$, and every $\vc{z}\in \mathcal{Z}$, 
\begin{align}
    0 \leq \ell(\vc{z},\hat{\vc{w}}) < \frac{2}{5n}+\frac{1}{4n^2} \eqqcolon 2\sigma. \label{def:sigmaproof}
\end{align}

Let $q_{\hat{\vc{W}}|S}\coloneqq q_{\hat{\vc{W}}}$ be the marginal distribution of $P_S p_{\hat{\vc{W}}|S}$. Now, part i. of Theorem~\ref{th:expectation} with the above choices of $p_{\hat{\vc{W}}|S}$ and $q_{\hat{\vc{W}}|S}$ and with $g(S,\hat{\vc{W}})\coloneqq  \gen(S,\hat{\vc{W}})$ and $\lambda \coloneqq n^2$, gives
\begin{align}
	\mathbb{E}_{S}[\gen(S,\vc{W}^{\{T\}})] \leq & \frac{1}{\lambda} \mathbb{E}_{P_S}\Big[D_{KL}\big(p_{\hat{\vc{W}}|S}\|q_{\hat{\vc{W}}}\big)\Big]+ \frac{1}{\lambda}\log\mathbb{E}_{P_S q_{\hat{\vc{W}}}}\Big[e^{\lambda \gen(S,\hat{\vc{W}})}\Big]+\epsilon \nonumber \\
 = & \frac{1}{\lambda} I(S;\hat{\vc{W}})+ \frac{1}{\lambda}\log\mathbb{E}_{P_S q_{\hat{\vc{W}}}}\Big[e^{\lambda \gen(S,\hat{\vc{W}})}\Big]+\epsilon \nonumber \\
 \leq &\frac{1}{\lambda}  I(S;\hat{\vc{W}})+ \frac{\lambda \sigma^2}{2n} +\epsilon \nonumber\\
 \leq &\frac{1}{\lambda}  I(S;\hat{\vc{W}})+ \frac{\lambda}{10n^3} +\epsilon \nonumber\\
 \leq  &\frac{1}{\lambda}  \left(H(\mathcal{E})+d \, h_b\big((1-r) 2^{-n})\big)\right)+ \frac{\lambda}{10n^3} +\epsilon  \nonumber\\
 \leq  &\frac{1}{\lambda}  \left(\frac{1}{18}  T \log_2(e) e^{-T/36} + \frac{3}{4} (1-r) T (n+1-\log_2(1-r))\right)+ \frac{\lambda}{10n^3} +\epsilon \nonumber \\
 =  &  \left(\frac{1}{9} \log_2(e) e^{-T/36} + \frac{3}{2} (1-r) (n+1-\log_2(1-r))\right)+ \frac{1}{10n} +\epsilon  \nonumber\\
 =  &  \mathcal{O}\left(\frac{1}{n}\right) +\epsilon 
 ,\label{eq:Limitation1}
\end{align}
where $h_b(x)=-x\log_2(x)-(1-x)\log_2(1-x)$ is the binary Shannon entropy that satisfies the inequality $h_b(x)\leq x(1-\log_2(x))$ and
\begin{align}
    \epsilon =\mathbb{E}_{S,\hat{\vc{W}}}\left[\gen(S,\vc{W}^{\{T\}})-\gen(S,\hat{\vc{W}})\right].
\end{align}
Hence, it remains to show that $\epsilon=\mathcal{O}\left(1/n\right)$. Re-write the distortion term as
\begin{align}
    \mathbb{E}_S\left[\gen\left(S,\vc{W}^{\{T\}}\right)-\gen(S,\hat{\vc{W}})\right] =  &\mathbb{E}_S\left[\mathcal{L}\left(\vc{W}^{\{T\}}\right)-\mathcal{L}(\hat{\vc{W}})\right] + \mathbb{E}_S\left[\hat{\mathcal{L}}(S,\hat{\vc{W}})-\hat{\mathcal{L}}\left(S,\vc{W}^{\{T\}}\right)\right]. \label{eq:distortionDecomp}
\end{align}
Now, we bound each term separately by $\mathcal{O}(1/n)$ which completes the proof.  Denote $\hat{\vc{W}}=(\hat{W}_1,\ldots,\hat{W}_d)$.
\begin{align}
  \mathbb{E}_S\left[\mathcal{L}\left(\vc{W}^{\{T\}}\right)-\mathcal{L}(\hat{\vc{W}})\right]  \leq &  \mathbb{P}\left(\mathcal{E}=0 \right)  \mathbb{E}_S\left[\mathcal{L}\left(\vc{W}^{\{T\}}\right)-\mathcal{L}(\hat{\vc{W}}) \bigg| \mathcal{E}=0 \right] + \mathbb{P}\left(\mathcal{E}=1\right)  \mathbb{E}_S\left[\mathcal{L}\left(\vc{W}^{\{T\}}\right)-\mathcal{L}(\hat{\vc{W}}) \bigg| \mathcal{E}=1  \right] \nonumber \\
  \stackrel{(a)}{\leq} &  \mathbb{P}\left(\mathcal{E}=0\right)  \mathbb{E}_S\left[\mathcal{L}\left(\vc{W}^{\{T\}}\right)-\mathcal{L}(\hat{\vc{W}}) \bigg| \mathcal{E}=0 \right] + (4+2/n)e^{-T/36} \nonumber\\
  \leq  &  \mathbb{E}_S\left[\mathcal{L}\left(\vc{W}^{\{T\}}\right)-\mathcal{L}(\hat{\vc{W}}) \bigg| \mathcal{E}=0 \right] + (4+2/n)e^{-T/36} \nonumber \\
  =  &  \mathbb{E}_S\left[\mathbb{E}_{\vc{Z}\sim \mu}\left[\ell(\vc{Z},\vc{W}^{\{T\}})-\ell(\vc{Z},\hat{\vc{W}})\right] \bigg| \mathcal{E}=0 \right] + (4+2/n)e^{-T/36} \nonumber\\
  \stackrel{(b)}{\leq} &  \frac{1}{2}\sum_{j\in[d]\setminus \mathcal{B}} \mathbb{E}_S\left[ (W_j^{\{T\}}-\hat{W}_{j})(W_j^{\{T\}}+\hat{W}_{j}+\lambda)\right] \nonumber\\
  &+ \frac{1}{2}\sum_{j\in \mathcal{B}} \mathbb{E}_S\left[ (W_j^{\{T\}}-\hat{W}_{j})(W_j^{\{T\}}+\hat{W}_{j}+\lambda)\right]+ (4+2/n)e^{-T/36} \nonumber\\
  \stackrel{(c)}{\leq} &  \frac{\lambda^2}{4}\sum_{j\in[d]\setminus \mathcal{B}} \mathbb{E}_S\left[ \left|\left(\left(1-2\eta \hat{\mu}_j\right)^T-\left(1-\eta \right)^T\right)\right|\right]+ \frac{1}{2}\sum_{j\in \mathcal{B}} \eta^2+ (4+2/n)e^{-T/36} \nonumber\\
  \leq&  \frac{\eta \lambda^2(T+1)}{2}\sum_{j\in[d]\setminus \mathcal{B}} \mathbb{E}_S\left[ \left|\hat{\mu}_j-\frac{1}{2}\right|\right] + \frac{\eta^2}{2} \|\vc{B}\|_0 + (4+2/n)e^{-T/36}\nonumber \\ 
  \stackrel{(d)}{\leq} &  \frac{\eta \lambda^2(T+1)}{4\sqrt{n}} (d-\|\vc{B}\|_0) \ + \frac{\eta^2}{2} \|\vc{B}\|_0 + (4+2/n)e^{-T/36} \nonumber\\
  =& \mathcal{O}\left(\frac{1}{n}\right),
\end{align}
where 
\begin{itemize}
    \item $(a)$ is concluded by using \eqref{eq:badCnumbers} and noting that $\ell(\vc{z},\vc{0})=0$ and $\ell(\vc{z},\vc{w}) \in [0,2+1/n]$ for every $\vc{z}\in \mathcal{Z}$ and $\vc{w} \in \mathcal{W}$,
    \item $(b)$ is derived by using Lemma~\ref{lem:GDdynamics}, by definition of the loss function $\ell(\vc{z},\vc{w})$, and by noting that the distortion is maximized for $r=1$,
    \item $(c)$ is obtained by Lemma~\ref{lem:GDdynamics},
    \item and $(d)$ by noting that $\mathbb{E}_S\left[ \left|\hat{\mu}_j-\frac{1}{2}\right|\right] \leq \frac{1}{2\sqrt{n}}$.    
\end{itemize}
Finally, for the second term of \eqref{eq:distortionDecomp} we have
\begin{align}
   \mathbb{E}_S&\left[\hat{\mathcal{L}}(S,\hat{\vc{W}})-\hat{\mathcal{L}}\left(S,\vc{W}^{\{T\}}\right)\right]\nonumber\\
   \leq &  \mathbb{P}\left(\mathcal{E}=0 \right)  \mathbb{E}_S\left[\hat{\mathcal{L}}(S,\hat{\vc{W}})-\hat{\mathcal{L}}\left(S,\vc{W}^{\{T\}}\right) \bigg| \mathcal{E}=0 \right] + \mathbb{P}\left(\mathcal{E}=1\right)  \mathbb{E}_S\left[\hat{\mathcal{L}}(S,\hat{\vc{W}})-\hat{\mathcal{L}}\left(S,\vc{W}^{\{T\}}\right) \bigg| \mathcal{E}=1  \right] \nonumber\\
  \leq &  \mathbb{P}\left(\mathcal{E}=0 \right)  \mathbb{E}_S\left[\hat{\mathcal{L}}(S,\hat{\vc{W}})-\hat{\mathcal{L}}\left(S,\vc{W}^{\{T\}}\right) \bigg| \mathcal{E}=0 \right] + (4+2/n)e^{-T/36} \nonumber\\
  \leq  &  \mathbb{E}_S\left[\hat{\mathcal{L}}(S,\hat{\vc{W}})-\hat{\mathcal{L}}\left(S,\vc{W}^{\{T\}}\right) \bigg| \mathcal{E}=0 \right] + (4+2/n)e^{-T/36} \nonumber\\
  =  &  \mathbb{E}_S\left[  \sum_{j\in[d]\setminus B} \hat{\mu}_j \left((\hat{W}_j-W^{\{T\}}_j\right)\left(\hat{W}_j+W^{\{T\}}_j+ \lambda 
  \right)  \bigg| \mathcal{E}=0 \right] + (4+2/n)e^{-T/36} \nonumber\\
  \leq  & \sum_{j\in[d]\setminus B}  \mathbb{E}_S\left[ \left| \left((\hat{W}_j-W^{\{T\}}_j\right)\left(\hat{W}_j+W^{\{T\}}_j+ \lambda 
  \right) \right|  \bigg| \mathcal{E}=0 \right] + (4+2/n)e^{-T/36} \nonumber\\
  \leq &  \frac{\lambda^2}{2}\sum_{j\in[d]\setminus \mathcal{B}} \mathbb{E}_S\left[ \left|\left(\left(1-2\eta \hat{\mu}_j\right)^T-\left(1-\eta \right)^T\right)\right|\right]+ (4+2/n)e^{-T/36} \nonumber\\
  \leq&  \eta \lambda^2(T+1)\sum_{j\in[d]\setminus \mathcal{B}} \mathbb{E}_S\left[ \left|\hat{\mu}_j-\frac{1}{2}\right|\right] + (4+2/n)e^{-T/36} \nonumber\\ 
  \leq &  \frac{\eta \lambda^2(T+1)}{2\sqrt{n}} (d-\|\vc{B}\|_0) \ + (4+2/n)e^{-T/36} \nonumber\\
  =& \mathcal{O}\left(\frac{1}{n}\right).
\end{align}

\subsubsection{Proof of~\texorpdfstring{\eqref{tail-bound-generalization-error-counter-example}}{(39)}}

The proof of part ii. is similar to the first part, but using Proposition~\ref{prop:PAClossy} instead of Theorem~\ref{th:expectation}. In Proposition~\ref{prop:PAClossy}, let $f(S,\vc{W}) \mapsto \lambda \gen(S,\vc{W}^{\{T\}})$, $g(S,\hat{\vc{W}})  \mapsto \lambda \gen(S,\hat{\vc{W}})$, $\epsilon \mapsto \lambda \epsilon$. Moreover, let $\pi$ be a deterministic distribution induced by GD. With these choices, with probability at least $1-\delta$,
\begin{align}
    \gen(S,\vc{W}^{\{T\}}) \leq \frac{1}{\lambda} D_{KL}\left(p_{\hat{\vc{W}}|S} \|q_{\hat{\vc{W}}|S}\right)+ \frac{1}{\lambda}\log\mathbb{E}_{P_S q_{\hat{\vc{W}}|S}}  \left[e^{ \lambda \gen(S,\hat{\vc{W}})}\right]+\frac{1}{\lambda} \log{(1/\delta)}+\epsilon. \label{eq:limitationPAC}
\end{align}
for any $q_{\hat{\vc{W}}|S}$ and any $p_{\hat{\vc{W}}|S}$ that should satisfy $\mathbb{E}_{p_{\hat{\vc{W}}|S}}\big[\gen(S,\vc{W}^{\{T\}})-\gen(S,\hat{\vc{W}})\big] \leq \epsilon$.

Now, we define $p_{\hat{\vc{W}}|S}$ and $q_{\hat{\vc{W}}|S}$. For a given $S$, if $\mathcal{E}=1$, then let $\hat{\vc{W}}=\vc{W}$ with probability one. In this case, $\epsilon=0$. Otherwise, if $\mathcal{E}=0$, choose the quantization rule as in part i, \ie  fix a given $r\in[1-\frac{1}{n^2},1]$ and for each $j\in[d]$, if $\hat{\mu}_j=0$, then let
\begin{align}
    \hat{W}_j = \begin{cases} v_0, & \text{with probability $r$}, \\
    v_1, & \text{with probability $1-r$},    
    \end{cases}
\end{align}
and if $\hat{\mu}_j>0$, then let $\hat{W}_j = v_0$ with probability one. Similar to part i. (but without taking expectation with respect to $S$), it can be shown that $\epsilon=\mathcal{O}(1/n)$ in \eqref{eq:limitationPAC}. 

If $\mathcal{E}=1$, let $q_{\hat{\vc{W}}|S}$ be equal to  a multi-variate Gaussian distribution $\mathcal{N}(0,\mathrm{I}_d)$ defined over $\mathcal{W}$, denoted as $q_{\hat{\vc{W}},1}$. Note that since $0\leq \ell(\vc{z},\vc{w}) \leq (2+1/n)$, hence, $\gen(S,\vc{W}) \leq (2+1/n)$.

If $\mathcal{E}=0$, let $q_{\hat{\vc{W}}|S}$ be a binary distribution, denoted by $q_{\hat{\vc{W}},2}$, taking value $v_1$ with probability $(1-r)2^{-n}\eqqcolon q$, and taking value $v_0$ with probability $(1-q)$.

It remains to show that whenever $\mathcal{E}=0$, then for some $\lambda$,
\begin{align}
    \frac{1}{\lambda} D_{KL}\left(p_{\hat{\vc{W}}|S} \|q_{\hat{\vc{W}}|S} \right)+ \frac{1}{\lambda}\log\mathbb{E}_{P_S q_{\hat{\vc{W}}|S}}  \left[e^{ \lambda \gen(S,\hat{\vc{W}})}\right]+\frac{1}{\lambda}\log(1/\delta) = \mathcal{O}(1/n). \label{eq:pacBayesLimiation1}
\end{align}

Whenever $\mathcal{E}=0$, we have
\begin{align*}
     D_{KL}\big(p_{\hat{\vc{W}}|S} \|q_{\hat{\vc{W}}|S} \big)+& \log\mathbb{E}_{P_S q_{\hat{\vc{W}}|S}}  \left[e^{ \lambda \gen(S,\hat{\vc{W}})}\right] \\
     =&D_{KL}\left(p_{\hat{\vc{W}}|S} \|q_{\hat{\vc{W}},2} \right)+ \log\mathbb{E}_{P_S q_{\hat{\vc{W}}|S}}  \left[e^{ \lambda \gen(S,\hat{\vc{W}})}\right] \\
     =&D_{KL}\left(p_{\hat{\vc{W}}|S} \|q_{\hat{\vc{W}},2} \right)+ \log\left(\mathbb{E}_{P_S q_{\hat{\vc{W}},2}}  \left[\mathbbm{1}_{\{\mathcal{E}=0\}}e^{ \lambda \gen(S,\hat{\vc{W}})}\right]+\mathbb{E}_{P_S q_{\hat{\vc{W}},1}}  \left[\mathbbm{1}_{\{\mathcal{E}=1\}}e^{ \lambda \gen(S,\hat{\vc{W}})}\right]\right) \\
     \leq &D_{KL}\left(p_{\hat{\vc{W}}|S} \|q_{\hat{\vc{W}},2} \right)+ \log\left(\mathbb{E}_{P_S q_{\hat{\vc{W}},2}}  \left[e^{ \lambda \gen(S,\hat{\vc{W}})}\right]+2e^{\lambda (2+1/n)}e^{-T/36}\right) \\
     \leq &  D_{KL}\left(p_{\hat{\vc{W}}|S} \|q_{\hat{\vc{W}}} \right)+ \log\left(e^{\frac{\lambda^2 \sigma^2}{2n}}+2e^{3\lambda-T/36}\right) \\
     \leq & \left(d-\|\vc{B}\|_0\right) \log\left(\frac{1}{1-q}\right) + \|\vc{B}\|_0 r \log\left(\frac{r}{1-q}\right)+ \|\vc{B}\|_0 (1-r) \log\left(\frac{1-r}{q}\right)\\
     &+ 2\max\left(\frac{\lambda^2 \sigma^2}{2n},3\lambda-T/36\right)+\mathcal{O}(1),
\end{align*}
where $\sigma$ is defined in \eqref{def:sigmaproof}. Plugging this in \eqref{eq:pacBayesLimiation1} and letting $\lambda =n^2/60$ give the desired result. Note that by considering only the case where $\mathcal{E}=0$, we showed \eqref{eq:pacBayesLimiation1} holds with probability $1-\delta-2e^{-T/36}$. Finally, observing that $2e^{-T/36}=o(1/n)$ terminates the proof of the equality.


\subsection{Proof of Lemma~\ref{lem:tailKL}} \label{pr:tailKL}
 Denote $\mathcal{B} \coloneqq \left\{(s,w) \in \supp(P_{S,W}) \colon f(s,w)> \Delta(s,w)\right\}$. If $P_{S,W}(\mathcal{B})=0$, then the lemma is proved. Assume then $P_{S,W}(\mathcal{B})>0$. Consider the distribution $\nu_{S,W}$ such that for any $(s,w) \in \mathcal{B}$, $\nu_{S,W}(s,w)\coloneqq P_{S,W}(s,w)/P_{S,W}(\mathcal{B})$, and otherwise $\nu(s,w)\coloneqq 0$. Note that by definition $\nu_{S,W} \in \mathcal{S}_{S,W}\left(f(s,w)- \Delta(s,w)\right)$.  Now, if $\nu_{S,W} \notin \mathcal{G}_{S,W}^{\delta}$, then $D_{KL}(\nu_{S,W}\|P_{S,W}) \geq \log(1/\delta)$. Hence, 
	\begin{align*}
		\log(1/\delta)\leq D_{KL}(\nu_{S,W}\|P_{S,W})=-\log(P_{S,W}(\mathcal{B}))=-\log \mathbb{P}\left(f(S,W) > \Delta(S,W) \right). 
	\end{align*}
	Thus, $\log \mathbb{P}\left(f(S,W) > \Delta(S,W) \right) \leq \log(\delta)$. This completes the proof of Lemma. 
	
	Otherwise, suppose that $\nu_{S,W} \in \mathcal{G}_{S,W}^{\delta}$, which yields $\nu_{S,W} \in \mathcal{F}_{S,W}^{\delta}$, due to the fact that $\nu_{S,W} \in \mathcal{S}_{S,W}\left(f(s,w)- \Delta(s,w)\right)$. Then, for this distribution and any $p_{\hat{W}|S}\in \mathcal{Q}(\nu_{S,W})$ we have
	\begin{align}
		\mathbb{E}_{\nu_{S,W} p_{\hat{W}|S}}\left[\Delta(S,W)-g(S,\hat{W})\right] \leq  \epsilon. \label{eq:lemtailKL1}
	\end{align}  
	Hence, it is optimal to let $\lambda=0$ in \eqref{eq:tailKL} for this distribution. Thus,
	\begin{align*}
		\inf_{p_{\hat{W}|S}\in \mathcal{Q}(\nu_{S,W})}\inf_{\lambda \geq 0}\left\{-D_{KL}(\nu_{S,W}\|P_{S,W})-\lambda \left(\mathbb{E}_{\nu_{S,W}}[\Delta(S,W)]-\epsilon-\mathbb{E}_{\nu_S P_{\hat{W}|S}} \left[g(S,\hat{W})\right] \right)\right\}\\
		=-D_{KL}(\nu_{S,W}\|P_{S,W})=\log(P_{S,W}(\mathcal{B}))=\log \mathbb{P}\left(f(S,W) > \Delta(S,W) \right).
	\end{align*} 
	This proves the inequality.


 \subsection{Proof of Lemma~\ref{lem:tailRenyi} } \label{pr:tailRenyi}
The proof is similar to the proof of Lemma~\ref{lem:tailKL}. Here, we repeat it for the sake of completeness.
	
	Denote $\mathcal{B} \coloneqq \left\{(s,w) \in \supp(P_{S,W}) \colon f(s,w)> \Delta(s,w)\right\}$. If $P_{S,W}(\mathcal{B})=0$, then the lemma is proved. Assume then $P_{S,W}(\mathcal{B})>0$. Consider the distribution $\nu_{S,W}$ such that for any $(s,w) \in \mathcal{B}$, $\nu_{S,W}(s,w)\coloneqq P_{S,W}(s,w)/P_{S,W}(\mathcal{B})$, and otherwise $\nu(s,w)\coloneqq 0$. Note that by definition $\nu_{S,W} \in \mathcal{S}_{S,W}\left(f(s,w)- \Delta(s,w)\right)$.  Now, if $\nu_{S,W} \notin \mathcal{G}_{S,W}^{\delta}$, then $D_{KL}(\nu_{S,W}\|P_{S,W}) \geq \log(1/\delta)$. Hence, 
	\begin{align*}
		\log(1/\delta)\leq D_{KL}(\nu_{S,W}\|P_{S,W})=-\log(P_{S,W}(\mathcal{B}))=-\log \mathbb{P}\left(f(S,W) > \Delta(S,W) \right). 
	\end{align*}
	Thus, $\log \mathbb{P}\left(f(S,W) > \Delta(S,W) \right) \leq \log(\delta)$. This completes the proof of Lemma. 
	
	Otherwise, suppose that $\nu_{S,W} \in \mathcal{G}_{S,W}^{\delta}$, which yields $\nu_{S,W} \in \mathcal{F}_{S,W}^{\delta}$, due to the fact that $\nu_{S,W} \in \mathcal{S}_{S,W}\left(f(s,w)- \Delta(s,w)\right)$. Then, for this distribution we have
	\begin{align}
		\mathbb{E}_{\nu_{S}}\left[\log \mathbb{E}_{\nu_{W|S}}[\Delta(S,W)]-\log\mathbb{E}_{\nu_{W|S}}\left[f(S,W)\right] \right] \leq  0. \label{eq:lemtailKL2}
	\end{align}  
	Hence, it is optimal to let $\lambda=0$ in \eqref{eq:tailKL} for this distribution. Thus,
	\begin{align*}
		\inf_{\lambda \geq 0}\left\{-D_{KL}(\nu_{S,W}\|P_{S,W})-\lambda \mathbb{E}_{\nu_{S}}\left[\log \mathbb{E}_{\nu_{W|S}}[\Delta(S,W)]-\log\mathbb{E}_{\nu_{W|S}}\left[f(S,W)\right] \right] \right\}\\
		=-D_{KL}(\nu_{S,W}\|P_{S,W})=\log(P_{S,W}(\mathcal{B}))=\log \mathbb{P}\left(f(S,W) > \Delta(S,W) \right).
	\end{align*} 
	This proves the inequality.

\end{document}